\pdfoutput=1

\documentclass[11pt]{article}

\usepackage[final]{acl}

\usepackage{times}
\usepackage{latexsym}

\usepackage[T1]{fontenc}

\usepackage[utf8]{inputenc}

\usepackage{microtype}

\usepackage{amssymb}
\usepackage{multirow}
\usepackage{hyperref}
\usepackage{url}
\usepackage{arydshln}
\usepackage{wrapfig}
\usepackage{amsmath}
\usepackage{colortbl}
\usepackage{xcolor}

\usepackage{enumitem}
\usepackage{tabularx}
\usepackage{adjustbox}
\usepackage{booktabs}

\usepackage{tcolorbox}
\usepackage{pgfplots}
\usepackage{makecell}
\usepackage{color}
\usepackage{circledtext}
\usepackage{textcomp}
\usepackage{epigraph}

\usepackage{footmisc}

\usetikzlibrary{shapes}
\usetikzlibrary{arrows,decorations.pathmorphing,backgrounds,positioning,fit,petri}
\usetikzlibrary{arrows.meta,fit,shapes.arrows}
\usetikzlibrary{positioning,shadows,patterns}
\usepackage{caption}
\usepackage{bm}

\definecolor{tiffanyblue}{RGB}{129,216,208}
\definecolor{bangdiblue}{RGB}{0,149,182}
\definecolor{kleinblue}{RGB}{0,47,167}

\usepackage{tikz}
\usepackage{subfig}
\usepackage{tkz-kiviat,pgfplots}
\usepackage{pgf-pie}

\usepackage{algorithm}

\usepackage{listings}
\usepackage{algpseudocode}

\pgfplotsset{compat=newest}
\usepgfplotslibrary{patchplots} 
\usepgfplotslibrary{fillbetween} 

\newtcolorbox{promptbox}[2][]{
	width=\columnwidth,
	colback = gray!8, 
	colframe = gray!8, 
	boxsep=0pt,left=0pt,right=10pt,top=0pt,bottom=0pt,
	title=#2,#1,
  fontupper=\footnotesize,
  before skip=4pt,
  after skip=4pt,
  }

\definecolor{myGreen}{RGB}{50, 168, 82}
\definecolor{myYellow}{RGB}{214, 169, 45}
\definecolor{myRed}{RGB}{201, 24, 24}
\definecolor{myBlue}{RGB}{50, 200, 200}
\definecolor{myBrown}{RGB}{140, 50, 50}
\definecolor{myPurple}{RGB}{128, 0, 200} 

\definecolor{given}{RGB}{197,217,197}
\definecolor{response}{RGB}{176,224,230}

\definecolor{color1}{RGB}{197,217,197}
\definecolor{color1_2}{RGB}{60,100,60}
\definecolor{color2}{RGB}{240,230,140}
\definecolor{color2_2}{RGB}{190,170,90}
\definecolor{color3}{RGB}{225,179,191}
\definecolor{color3_2}{RGB}{175,129,141}
\definecolor{color4}{RGB}{176,224,230}

\title{\textbf{\textit{x1}}: Learning to Think Adaptively Across Languages and Cultures}

\author{
  Yangfan Ye$^{1}$,
  Xiaocheng Feng$^{1,2}$\thanks{Corresponding Author},
  Xiachong Feng$^{3}$,
  Yichong Huang$^{1}$,
  Zekun Yuan$^{1}$,\\
  \textbf{Lei Huang$^{1}$ 
  Weitao Ma$^{1}$,
  Qichen Hong$^{4}$,
  Yunfei Lu$^{4}$,
  Dandan Tu$^{4}$,
  Bing Qin$^{1,2}$} \\
  $^{1}$Harbin Institute of Technology \quad
  $^{2}$Peng Cheng Laboratory \\
  $^{3}$The University of Hong Kong \quad
  $^{4}$Huawei Technologies Co., Ltd \\
  \texttt{\{yfye,xcfeng\}@ir.hit.edu.cn}\\
}

\begin{document}
\maketitle

\begin{abstract}
Languages encode distinct abstractions and inductive priors, yet most large language models (LLMs) overlook this diversity by reasoning in a single dominant language.
In this work, we introduce \textit{x1}, a family of reasoning models that can \emph{adaptively} reason in an advantageous language on a per-instance basis.
To isolate the effect of reasoning-language choice, \textit{x1} is constructed without expanding the model's knowledge boundaries and is trained by contrasting linguistically distinct reasoning trajectories for the same input.
Our extensive experiments demonstrate the benefits of adaptive multilingual reasoning in multilingual mathematical and culturally grounded tasks.
Moreover, our results challenge a simplistic view of scaling law: while scaling reduces cross-lingual disparities in procedural domains such as math reasoning, it does not eliminate the advantages of reasoning in culture-associated languages in cultural scenarios, as we empirically show that such reasoning enables more efficient and accurate cultural knowledge recall.
Overall, our findings establish language choice as a functional component of reasoning, with implications for building more generalist and globally competent reasoning models.\footnote{\url{https://github.com/YYF-Tommy/x1-adaptive-multilingual-reasoning}}
\end{abstract}
\section{Introduction}\label{sec:intro}

\begin{flushleft}
\textit{\quad``The limits of my language mean \\ \quad\quad\quad\quad\quad\quad\quad\quad\quad the limits of my world.''} \\
\end{flushleft}
\begin{flushright}
\textit{--- Ludwig Wittgenstein, 1922~\citep{wittgenstein2023tractatus}}
\end{flushright}

``Reasoning'' endows large language models (LLMs) with the ability to go beyond surface-level pattern matching and to tackle complex tasks such as competition-level mathematics, logical reasoning, and multi-hop question answering~\citep{wei2022chain,snell2024scaling,brown2024large}.
However, most existing LLMs, like OpenAI-o1/o3~\citep{jaech2024openai} and Qwen3~\citep{yang2025qwen3}, reason predominantly in high-resource languages (such as English, Chinese).
While reasoning in a dominant language is often effective, it implicitly assumes a single linguistic perspective to be universally optimal.
However, languages encode distinct abstractions, cultural priors, and modes of expression, which can subtly shape how problems are decomposed, interpreted, and solved~\citep{goddard2003thinking,kovecses2006language}.
Consequently, confining reasoning to a single language may underutilize this diversity, particularly in multilingual and culturally grounded scenarios.

\definecolor{my-red}{RGB}{239,138,150}      
\definecolor{my-yellow}{RGB}{253,190,133}   
\definecolor{my-blue}{RGB}{166,206,227}     
\definecolor{my-green}{RGB}{178,223,138}    

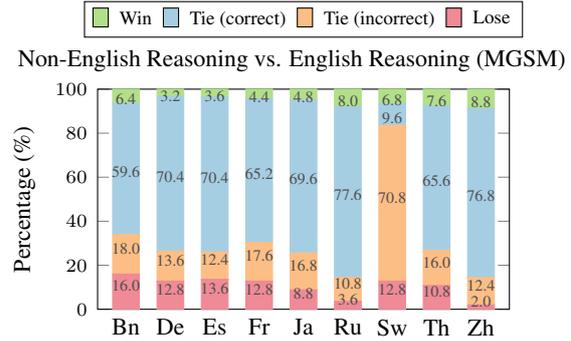
\begin{figure}[t]
    \centering
    \pgfplotsset{width=0.5\linewidth, height=1\linewidth, compat=1.18}
    \begin{tikzpicture}
        \begin{axis}[
            ybar stacked,
            bar width=10pt,
            width=7cm,
            height=4.5cm,
            ymin=0, ymax=100,
            ylabel={Percentage (\%)},
            ylabel style={font=\small}, 
            symbolic x coords={Bn, De, Es, Fr, Ja, Ru, Sw, Th, Zh},
            xtick=data,
            xticklabel style={font=\small},
            title={Non-English Reasoning vs. English Reasoning (MGSM)},
            title style={xshift=-0.4em, yshift=-0.3em, font=\small},
            tick label style={font=\small},
            enlarge x limits=0.08,
            yticklabel style={font=\scriptsize},
            xticklabel style={font=\footnotesize},
            legend style={
                reverse legend,
                font=\footnotesize,
                legend columns=4,
                column sep=0.5ex,
                nodes={scale=0.8, transform shape},
                at={(0.5,1.4)}, anchor=north
            },
            legend image code/.code={
                \draw[draw=black] 
                (0cm,-0.1cm) rectangle (0.2cm,0.15cm);
            },
        ]

        \addplot+[ybar, fill=my-red, draw=my-red,
          nodes near coords,
          point meta=explicit symbolic,
          every node near coord/.append style={font=\tiny, xshift=0ex, yshift=0.2em, text=black!70}] coordinates {
            (Bn,16.0) [16.0] (De,12.8) [12.8] (Es,13.6) [13.6] (Fr,12.8) [12.8] 
            (Ja,8.8) [8.8] (Ru,3.6) [3.6] (Sw,12.8) [12.8] 
            (Th,10.8) [10.8] (Zh,2.0) [2.0]
        };

        \addplot+[ybar, fill=my-yellow, draw=my-yellow,
          nodes near coords,
          point meta=explicit symbolic,
          every node near coord/.append style={font=\tiny, xshift=0ex, yshift=0.2em, text=black!70}] coordinates {
            (Bn,18.0) [18.0] (De,13.6) [13.6] (Es,12.4) [12.4] (Fr,17.6) [17.6] 
            (Ja,16.8) [16.8] (Ru,10.8) [10.8] (Sw,70.8) [70.8]
            (Th,16.0) [16.0] (Zh,12.4) [12.4]
        };

        \addplot+[ybar, fill=my-blue, draw=my-blue,
          nodes near coords,
          point meta=explicit symbolic,
          every node near coord/.append style={font=\tiny, xshift=0ex, yshift=-0.1em, text=black!70}] coordinates {
            (Bn,59.6) [59.6] (De,70.4) [70.4] (Es,70.4) [70.4] (Fr,65.2) [65.2] 
            (Ja,69.6) [69.6] (Ru,77.6) [77.6] (Sw,9.6) [9.6]
            (Th,65.6) [65.6] (Zh,76.8) [76.8]
        };

        \addplot+[ybar, fill=my-green, draw=my-green,
          nodes near coords,
          point meta=explicit symbolic,
          every node near coord/.append style={font=\tiny, xshift=0ex, yshift=-0.1em, text=black!70}] coordinates {
            (Bn,6.4) [6.4] (De,3.2) [3.2] (Es,3.6) [3.6] (Fr,4.4) [4.4] 
            (Ja,4.8) [4.8] (Ru,8.0) [8.0] (Sw,6.8) [6.8]
            (Th,7.6) [7.6] (Zh,8.8) [8.8]
        };

        \legend{Lose, Tie (incorrect), Tie (correct), Win}
        \end{axis}
    \end{tikzpicture}
    \vspace{-0.3\baselineskip}
    \caption{Win/tie/lose rates of non-English vs. English reasoning pathways for \textit{Qwen3-4B} on \textit{MGSM}. Reasoning in contrast languages is enabled via Step~1 (introduced in \S\ref{sec:intro}). ``Win'': cases where non-English reasoning outperforms English reasoning; ``Tie'' (correct/incorrect): both reasoning pathways yield correct/incorrect answers; ``Lose'': cases where English reasoning outperforms non-English reasoning.}
    \vspace{-0.7\baselineskip}
    \label{fig:pilot}
\end{figure}

\begin{figure*}[t]
    \centering
    \includegraphics[width=1\textwidth]{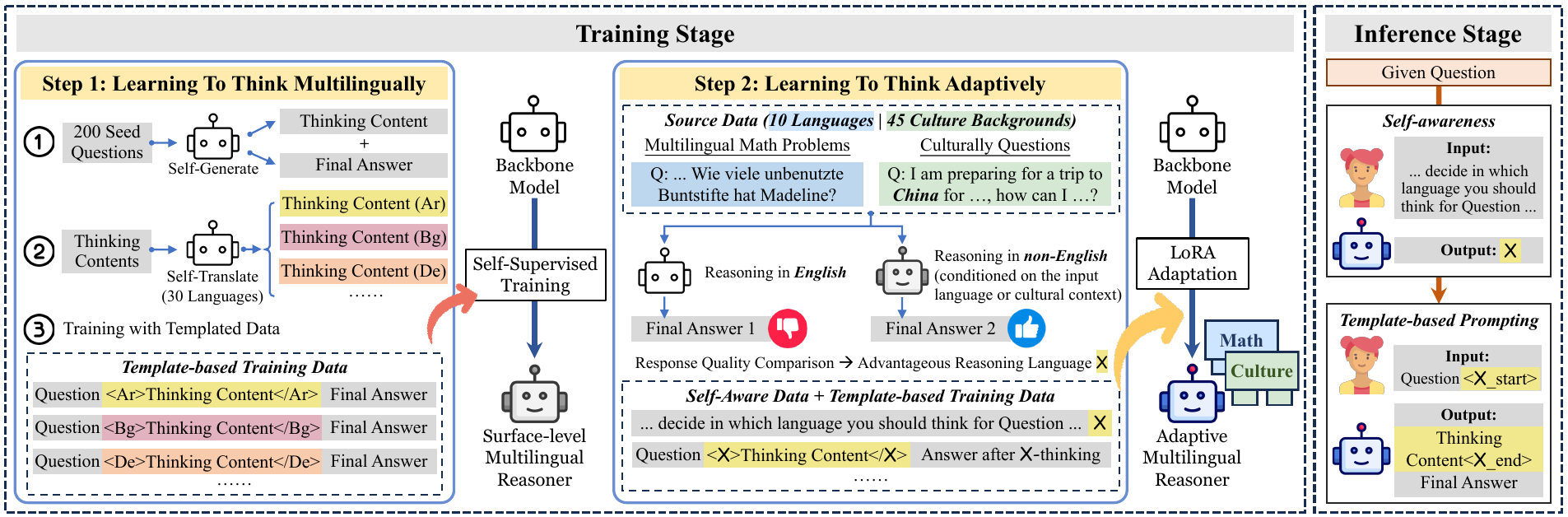}
    \caption{Overview of the construction process and inference usage of \textit{x1} models.
    }
    \vspace{-0.5\baselineskip}
    \label{fig:method}
\end{figure*}

Recently, \citet{yong2025crosslingual} conducted a detailed analysis of cross-lingual reasoning and demonstrate that, in multilingual mathematics problems, English is not always the optimal language for reasoning. In Figure~\ref{fig:pilot}, we quantify the advantages of non-English reasoning by measuring the win/tie/lose rates of non-English reasoning compared to English reasoning, revealing that in \textit{MGSM}~\citep{shi2022language}, roughly 3\%--9\% (varying across languages) of instances can be solved by non-English reasoning paths where English reasoning fails.
Beyond mathematical reasoning, \citet{yong2025crosslingual} observed that test-time scaling of English reasoning (i.e., extending the reasoning length) does not necessarily improve performance in culture-aware scenarios, where success may require more knowledge recall or culturally grounded reasoning.
This limitation further motivates our investigation into non-English reasoning pathways, particularly for culturally situated reasoning, where the choice of thinking language may play a decisive role.

In this paper, we take a step toward exploiting the diversity of reasoning perspectives offered by English and non-English languages.
Building on existing open-source reasoning models, we introduce \textbf{\textit{x1}}, a family of models that can \emph{adaptively select} the more advantageous thinking language on a per-instance basis.
Crucially, this is achieved \emph{without expanding the model's knowledge boundaries}, allowing us to isolate the effect of reasoning-language choice in a controlled and comparable setting, rather than emphasizing absolute performance gains.
Specifically, we first equip the model with the ability to reason in a \emph{specified language} across multiple languages through self-supervised training, and then induce adaptive reasoning behavior by contrasting linguistically distinct reasoning trajectories for the same input.

We investigate adaptive multilingual reasoning in two deeply language-intertwined scenarios: multilingual mathematical reasoning and cultural reasoning.
Our findings suggest that reasoning in LLMs is not language-agnostic, but shaped by the reasoning language choice.
By enabling adaptive multilingual reasoning, we show that linguistic diversity offers systematic benefits beyond reliance on a single language.
Importantly, our results challenge a simplistic view of scaling laws: while scaling reduces cross-lingual disparities in procedural domains such as mathematical reasoning, it fails to eliminate the intrinsic advantages of reasoning in culture-associated languages for culturally grounded tasks, which leads to more efficient and accurate cultural knowledge recall.
Together, these findings indicate that language choice is not merely a surface realization of thought, but a functional component of reasoning itself, with important implications for building truly generalist and globally competent reasoning models.

\section{Constructing \textit{x1}}\label{sec:method}
To achieve reasoning in a advantageous language without expanding the model's knowledge boundaries, we propose a two-stage training strategy that leverages self-generated contrastive signals to induce adaptive multilingual reasoning.

\subsection{Step 1: To be a Multilingual Reasoner}\label{sec:surface}

The first step aims to open up the model's multilingual reasoning space, enabling it to deliberately reason in a specified language rather than being implicitly anchored to its default thinking language.

To achieve this, we adopt a lightweight self-supervised training strategy that teaches the model to perform reasoning in a required language while ensuring that the final answer remains consistent with the prompting language\footnote{For example, if the prompt is in English but the reasoning language is constrained to German, the final response should still be generated in English.}.
In practice, we first query the backbone model $\mathcal{M}$ with 200 seed questions sampled from \textit{Flan\_v2}~\citep{wei2021finetuned}, collecting its generated reasoning traces $\{\operatorname{t}_i\}_{i=1}^{200}$ and final answers $\{\operatorname{a}_i\}_{i=1}^{200}$.
We then instruct $\mathcal{M}$ itself to translate each reasoning trace into 30 different languages $\{l_j\}_{j=1}^{30}$ (listed in Appendix~\ref{app:step1}), yielding a multilingual set of reasoning traces $\{\operatorname{t}_i^{l_j}\}_{i=1}^{200}$ and $j=1,\dots,30$.
To ensure linguistic fidelity, we apply quality filtering using COMET\footnote{Unbabel/wmt23-cometkiwi-da-xxl~\citep{rei2023scaling}} scores and discard translations of evidently poor quality.

Using these translated reasoning traces, we finetune the backbone model $\mathcal{M}$ with a templated format that explicitly specifies a reasoning language as follows:
(see complete examples in Figure~\ref{fig:step1_cases})

\begin{promptbox}
    
    // Example of reasoning in \textbf{Arabic} \\
	Input: {\setlength{\fboxsep}{0pt}\colorbox{color1}{\{$\operatorname{Question}_i$\}}}\\
	Output: <think>\verb|\|n<{\textcolor{color2_2}{\textbf{Arabic\_start}}}>\verb|\|n\verb|\|n{\setlength{\fboxsep}{0pt}\colorbox{color2}{$\{\operatorname{t}_i^{\operatorname{Ar}}$\}}}\verb|\|n\verb|\|n \\
    \textcolor{gray!8}{Output: }<{\textcolor{color2_2}{\textbf{Arabic\_end}}}>\verb|\|n</think>\verb|\|n\verb|\|n{\setlength{\fboxsep}{0pt}\colorbox{color1}{$\{\operatorname{a}_i$\}}}
\end{promptbox}

The resulting ``sibling'' model, $\mathcal{M}_{\operatorname{surface}}$, can be viewed as a \emph{surface-level multilingual reasoner}: it is capable of following simple templates to reason in a specified language \textit{$X$} by appending a simple template ``\textit{<think>\textbackslash n<$X$\_start>}'' at inference time, while avoiding undesirable language drift.
Notably, this capability operates purely at the level of reasoning expression: it exposes alternative language-conditioned reasoning paths and does not introduce any external knowledge or capability expansion (such as Google Translation, external knowledge bases, or stronger auxiliary models).

\subsection{Step 2: To be an Adaptive Reasoner}

The goal of Step~2 is to elevate multilingual reasoning from language following to adaptively choose the more advantageous reasoning language.

Leveraging $\mathcal{M}_{\operatorname{surface}}$, we are able to generate paired reasoning trajectories in English and non-English languages for each input\footnote{If $\mathcal{M}$ naturally reasons in English, we require $\mathcal{M}_{\operatorname{surface}}$ to reason in a non-English language; conversely, we require $\mathcal{M}_{\operatorname{surface}}$ to reason in English.\label{myfoot}}. These paired trajectories can be viewed as \emph{counterfactual reasoning paths} induced by different language choices, providing a basis for comparative learning.

\paragraph{Training Data Source.}
We study adaptive reasoning in two language-intertwined scenarios.
\textbf{(1) Multilingual Math Problems} ($D_{\operatorname{math}}$): We use the \textit{MGSM8KInstruct} dataset~\citep{chen2023breaking}, which contains math problems paired with their correct answers. We sample 200 samples across 10 languages, resulting in 2,000 training instances.
\textbf{(2) Culture-related Problems} ($D_{\operatorname{culture}}$): We use the \textit{CultureBank} dataset~\citep{shi2024culturebank}, which provides cultural questions along with their underlying cultural knowledge. We sample 100-200 samples from each of the 25 language groups (depending on availability), covering cultural questions from 45 countries/regions, resulting in 4,413 samples in total. (Detailed statistics are in Appendix~\ref{app:step2})

\paragraph{Advantageous Language Identification.}
For each question in $D_{\operatorname{math}}$ or $D_{\operatorname{culture}}$, we first obtain the paired reasoning trajectories in both English and non-English with $\mathcal{M}$ and $\mathcal{M}_{\operatorname{surface}}$\footref{myfoot}.
We identify the advantageous reasoning language based on the quality of the final answers, under the principle that stronger reasoning should manifest in superior outputs.
For multilingual math problems, answer quality is determined via exact numerical matching.
For culture-related questions, we employ \textit{LLM-as-a-Judge}~\citep{zheng2023judging} to score how well the response entails the relevant cultural knowledge.
The reasoning language associated with the higher-scoring trajectory is then regarded as the advantageous thinking language for that instance.
(Detailed implementations for comparison are provided in Appendix~\ref{app:step2}.)

\definecolor{up-green}{RGB}{0,120,0}
\definecolor{down-red}{RGB}{255,0,0}
\definecolor{ForestGreen}{RGB}{34,139,34}

\begin{table*}[t]
\small
\renewcommand{\arraystretch}{1.2} 
\setlength{\dashlinedash}{4pt} 
\setlength{\dashlinegap}{2pt}  
  \centering
  \resizebox{\linewidth}{!}{
    \begin{tabular}{l@{\hskip 4pt}cccccccc}
    \toprule
    \multirow{3}[4]{*}{\textbf{Models}} & \multicolumn{4}{c}{\textbf{Multilingual Math Reasoning}} & \multicolumn{4}{c}{\textbf{Cultural Reasoning}} \\
    \cmidrule(lr){2-5}
    \cmidrule(lr){6-9}
    & \multicolumn{2}{c}{\textbf{MGSM}} & \multicolumn{2}{c}{\textbf{MT-AIME}} & \multicolumn{2}{c}{\textbf{FORK}} & \multicolumn{2}{c}{\textbf{CulturalBench}} \\
    & \multicolumn{1}{c}{\textit{Non-Think}} & \multicolumn{1}{c}{\underline{\textbf{\textit{Think}}}} & \multicolumn{1}{c}{\textit{Non-Think}} & \multicolumn{1}{c}{\underline{\textbf{\textit{Think}}}} & \multicolumn{1}{c}{\textit{Non-Think}} & \multicolumn{1}{c}{\underline{\textbf{\textit{Think}}}} & \multicolumn{1}{c}{\textit{Non-Think}} & \multicolumn{1}{c}{\underline{\textbf{\textit{Think}}}} \\
    \midrule
    \multicolumn{9}{c}{\textbf{\textit{Top-tiered Reasoning Models}}} \\
    \midrule
    \text{o4-mini-high} & -- & 82.32 & -- & 75.33 & -- & 74.46 & -- & 82.31 \\
    \text{DeepSeek-V3.2} & 79.24 & 76.32 & 53.00 & 85.67 & 72.83 & 78.80 & 84.60 & 89.16 \\
    \midrule
    \multicolumn{9}{c}{\textbf{\textit{Open-sourced Reasoning Models}}} \\
    \midrule
    \text{Qwen3-4B} & 70.21 & 76.59 & 12.89 & 21.78 & 74.64 & 73.73 & 68.76 & 70.85 \\
    \text{Qwen3-14B} & 77.64 & 82.56 & 19.33 & 29.22 & 74.46 & 73.91 & 75.50 & 78.24 \\
    \text{Qwen3-32B} & 80.52 & 83.98 & 21.83 & 33.89 & 78.26 & 81.88 & 78.76 & 81.26 \\
    \text{DeepSeek-R1-Distill-Qwen-7B} & 54.76 & 60.05 & 8.33 & 25.83 & 40.04 & 55.25 & 29.58 & 38.63 \\
    \text{DeepSeek-R1-Distill-Llama-8B} & 38.17 & 40.36 & 2.67 & 14.44 & 37.86 & 74.09 & 35.02 & 57.19 \\
    \midrule
    \multicolumn{9}{c}{\textbf{\textit{x1 Series Models}}} \\
    \midrule
    \noalign{\vskip -0.8ex}
    & \multicolumn{4}{c}{\textbf{\textit{\cellcolor{cyan!15}+ Math}}} & \multicolumn{4}{c}{\textbf{\textit{\cellcolor{up-green!15}+ Culture}}} \\
    {\textbf{\textit{x1}}-Qwen3-4B}
    & 70.30 & 77.69 (\textcolor{ForestGreen}{$\bm{\uparrow}$} 1.10) & 13.56 & 22.83 (\textcolor{ForestGreen}{$\bm{\uparrow}$} 1.05) & 75.18 & 78.08 (\textcolor{ForestGreen}{$\bm{\uparrow}$} 4.35) & 68.46 & 72.74 (\textcolor{ForestGreen}{$\bm{\uparrow}$} 1.89) \\
    {\textbf{\textit{x1}}-Qwen3-14B}
    & 77.38 & 83.64 (\textcolor{ForestGreen}{$\bm{\uparrow}$} 1.08) & 19.44 & 33.11 (\textcolor{ForestGreen}{$\bm{\uparrow}$} 3.89) & 73.12 & 76.81 (\textcolor{ForestGreen}{$\bm{\uparrow}$} 2.90) & 76.07 & 81.58 (\textcolor{ForestGreen}{$\bm{\uparrow}$} 3.34) \\
    {\textbf{\textit{x1}}-Qwen3-32B}
    & 80.12 & 84.43 (\textcolor{ForestGreen}{$\bm{\uparrow}$} 0.45) & 22.11 & 34.50 (\textcolor{ForestGreen}{$\bm{\uparrow}$} 0.61) & 79.17 & 82.61 (\textcolor{ForestGreen}{$\bm{\uparrow}$} 0.73) & 80.47 & 81.99 (\textcolor{ForestGreen}{$\bm{\uparrow}$} 0.73) \\
    {\textbf{\textit{x1}}-DeepSeek-R1-Distill-Qwen-7B}
    & 54.52 & 63.24 (\textcolor{ForestGreen}{$\bm{\uparrow}$} 3.19) & 9.00 & 27.00 (\textcolor{ForestGreen}{$\bm{\uparrow}$} 1.17) & 54.89 & 58.70 (\textcolor{ForestGreen}{$\bm{\uparrow}$} 3.45) & 44.82 & 49.04 (\textcolor{ForestGreen}{$\bm{\uparrow}$} 10.4) \\
    {\textbf{\textit{x1}}-DeepSeek-R1-Distill-Llama-8B}
    & 38.01 & 52.17 (\textcolor{ForestGreen}{$\bm{\uparrow}$} 11.8) & 2.89 & 17.00 (\textcolor{ForestGreen}{$\bm{\uparrow}$} 2.56) & 57.07 & 76.27 (\textcolor{ForestGreen}{$\bm{\uparrow}$} 2.18) & 51.13 & 60.88 (\textcolor{ForestGreen}{$\bm{\uparrow}$} 3.69) \\
    \bottomrule
    \end{tabular}
  }
  \caption{Mean@3 results that are the averages of the performance across all languages / cultures for each dataset.}
  \vspace{-0.5\baselineskip}
  \label{tab:main}
\end{table*}

\paragraph{Training Data Construction.}

For each instance, we retain only the reasoning trace and answer corresponding to the advantageous language, and discard cases where English and non-English reasoning receive tied scores, to enforces a sharper contrast between alternative reasoning pathways and provides a clear signal for language preference.

We train the backbone model $\mathcal{M}$ using a templated format (similar to Step~1) that explicitly specifies the selected reasoning language \textit{$X$}, implemented via parameter-efficient LoRA finetuning.
For a given question $Q$, if language $X$ is identified as the advantageous reasoning language, with corresponding reasoning trace $T$ and final answer $A$, we construct the training instance as follows:
\begin{promptbox}
    
	Input: {\setlength{\fboxsep}{0pt}\colorbox{color1}{\{$\operatorname{Q}$\}}}\\
	Output: <think>\verb|\|n<{\textcolor{color2_2}{\textbf{X\_start}}}>\verb|\|n\verb|\|n{\setlength{\fboxsep}{0pt}\colorbox{color2}{$\{\operatorname{T}$\}}}\verb|\|n\verb|\|n<{\textcolor{color2_2}{\textbf{X\_end}}}>\verb|\|n \\
	\textcolor{gray!8}{Output: }</think>\verb|\|n\verb|\|n{\setlength{\fboxsep}{0pt}\colorbox{color1}{$\{\operatorname{A}$\}}}
\end{promptbox}

Additionally, we introduce auxiliary self-awareness data that externalizes the model's language-selection decision as an explicit prediction task.
This encourages the model to internalize reasoning-language choice as a deliberate component of its reasoning strategy rather than as a fixed or implicit heuristic. The data format is illustrated below (see complete example in Figure~\ref{fig:step2_cases}):
\begin{promptbox}
    
	Input: ...decide in which language you should internally \textcolor{gray!8}{Input: }think...for question {\setlength{\fboxsep}{0pt}\colorbox{color1}{\{$\operatorname{Q}$\}}}\verb|\|n\verb|\|nThinking Language:\\
	Output: <think>\verb|\|n\verb|\|n</think>\verb|\|n\verb|\|n{\setlength{\fboxsep}{0pt}\colorbox{color1}{$\{\mathit{X}$\}}}
\end{promptbox}

All training data in Step~2 are self-generated by either $\mathcal{M}$ or $\mathcal{M}_{\operatorname{surface}}$.
This stage relies on instance-level comparisons between alternative reasoning-language choices, enabling the backbone models to select and exploit its latent multilingual reasoning capabilities in an adaptive manner.

\section{Experiments}\label{sec:exp}

\subsection{Setup}

\paragraph{Models.}
We perform training on five representative reasoning models. Qwen3-series models~\citep{yang2025qwen3}: (1) \textit{Qwen3-4B}, (2) \textit{Qwen3-14B}, (3) \textit{Qwen3-32B} and DeepSeek-R1-distilled models~\citep{guo2025deepseek}: (4) \textit{DeepSeek-R1-Distill-Qwen-7B} and (5) \textit{DeepSeek-R1-Distill-Llama-8B}.

\paragraph{Training.}
Our training processes are conducted on \textit{8$\times$A800-SXM4-80GB} GPUs.
In step-1 self-supervised training, we perform full-parameter finetuning~\citep{ouyang2022training} for all models to ensure effective acquisition of the designed templates, except for \textit{Qwen3-32B}, for which we adopt LoRA~\citep{hu2022lora} training due to its large model size.
In step-2, we apply LoRA training separately for multilingual math and cultural reasoning scenarios,
aiming to minimally modify the backbone models while enabling lightweight and flexible scenario adaptation.
More training details and hyperparameters are provided in Appendix~\ref{app:exp}.

\paragraph{Evaluations.}
We conduct experiments on four benchmarks, which can be categorized into:
\begin{itemize}[leftmargin=*]
\setlength{\parsep}{0pt}
\setlength{\parskip}{0pt}
\item \textbf{Multilingual Math Reasoning:} (1) \textit{MGSM}~\citep{shi2022language}, the multilingual version of grade-school math problems translated by human annotators and (2) \textit{MT-AIME}~\citep{son2025linguistic}, the multilingual version of challenging mathematical problems from American Invitational Mathematics Examination (AIME). For multilingual math reasoning, our main experiments cover 10 languages: \textit{Bn, De, En, Es, Fr, Ja, Ru, Sw, Th, Zh}.
\item \textbf{Cultural Reasoning:} (1) \textit{FORK}~\citep{palta-rudinger-2023-fork}, a manually-curated set of English questions for probing cultural biases present in commonsense reasoning, with a specific focus on food-related customs, covering 10 global regions and (2) \textit{CulturalBench}~\citep{chiu-etal-2025-culturalbench}, a set of 1,696 human written and human-verified English questions to assess LMs' cultural knowledge, covering 45 global regions.
\end{itemize}
We follow the default official sampling parameters for each model and we report the Mean@3 (average results of three runs) performance in our main experiments. 
We set the maximum number of new tokens to 32,768 to ensure sufficient output length.
To prevent meaningless repetitive reasoning during inference, we adopt a repetition-detection truncation mechanism to reduce unnecessary token generation and computational overhead. The implementations are provided in the Appendix~\ref{app:truncation}.

\subsection{Overall Performance}

The average Mean@3 results over different languages / cultures are presented in Table~\ref{tab:main}.
The results for each language are in \S\ref{app:subset_exp} Table~\ref{tab:main_math}.
We also compare and anaylze the performance of \textit{vanilla finetuning} and \textit{majority voting} in Appendix~\S\ref{app:supp_exp}.

\begin{table*}[t]
    \small
    \renewcommand{\arraystretch}{1.2} 
    \setlength{\tabcolsep}{4pt}
    \setlength{\dashlinedash}{4pt} 
    \setlength{\dashlinegap}{2pt}  
      \centering
      \resizebox{\textwidth}{!}{
        \begin{tabular}{rcccccccccc}
        \toprule
        \multicolumn{11}{c}{\textit{Data Format: (Thinking Phase | Answer Phase | Both Phases)}} \\
        \midrule
        \multicolumn{11}{c}{Language Compliance Rate (\%) on MGSM} \\
        \cmidrule(lr){1-11}
        Query Language& Bn & De & En & Es & Fr & Ja & Ru & Sw & Th & Zh \\
        \midrule
        \text{Qwen3-4B} &  --~~ | 95.6 | ~~-- & --~~ | 96.8 | ~~-- & --~~ | 100 | ~~-- & --~~ | 98.0 | ~~-- & --~~ | 99.2 | ~~-- & --~~ | 96.8 | ~~-- & --~~ | 99.6 | ~~-- & --~~ | 26.4 | ~~-- & --~~ | 92.4 | ~~-- & --~~ | 100 | ~~-- \\
        \text{\textit{x1}-Qwen3-4B} & 95.6 | 97.6 | 93.2 & 100 | 99.2 | 99.2 & 100 | 100 | 100 & 99.6 | 99.2 | 98.8 & 99.2 | 99.6 | 98.8 & 100 | 99.6 | 99.6 & 99.6 | 100 | 99.6 & 99.6 | 24.0 | 23.6 & 98.0 | 95.6 | 94.0 & 100 | 99.6 | 99.6\\
        \cdashline{1-11}\noalign{\vskip 0.4ex}
        \text{Qwen3-14B} &  --~~ | 97.6 | ~~-- & --~~ | 97.2 | ~~-- & --~~ | 100 | ~~-- & --~~ | 92.0 | ~~-- & --~~ | 89.6 | ~~-- & --~~ | 99.6 | ~~-- & --~~ | 99.6 | ~~-- & --~~ | 49.6 | ~~-- & --~~ | 99.2 | ~~-- & --~~ | 99.6 | ~~-- \\
        \text{\textit{x1}-Qwen3-14B} & 99.6 | 70.8 | 70.40 & 99.2 | 93.6 | 92.8 & 100 | 99.6 | 99.6 & 100 | 92.0 | 92.0 & 99.2 | 92.4 | 91.6 & 100 | 97.2 | 97.2 & 92.0 | 96.0 | 88.0 & 100 | 54.4 | 54.4 & 98.8 | 94.0 | 92.8 & 100 | 100 | 100 \\
        \cdashline{1-11}\noalign{\vskip 0.4ex}
        \text{Qwen3-32B} &  --~~ | 93.6 | ~~-- & --~~ | 94.8 | ~~-- & --~~ | 100 | ~~-- & --~~ | 93.6 | ~~-- & --~~ | 96.0 | ~~-- & --~~ | 91.2 | ~~-- & --~~ | 99.6 | ~~-- & --~~ | 64.0 | ~~-- & --~~ | 92.4 | ~~-- & --~~ | 100 | ~~-- \\
        \text{\textit{x1}-Qwen3-32B} & 98.8 | 90.0 | 89.2 & 93.2 | 93.2 | 86.4 & 100 | 100 | 100 & 100 | 90.8 | 90.8 & 100 | 96.0 | 96.0 & 100 | 90.0 | 90.0 & 98.8 | 93.6 | 92.4 & 99.6 | 64.8 | 64.8 & 98.8 | 86.8 | 86.4 & 99.6 | 100 | 99.6 \\
        \bottomrule
        \end{tabular}
      }
      \vspace{-0.2\baselineskip}
      \caption{Language compliance results of \textit{x1} models and their backbones on \textit{MGSM} across different languages.}
      \vspace{-0.8\baselineskip}
      \label{tab:lang_compliance}
    \end{table*}


\paragraph{(1) Evidence of Decoupling in Math and Cultural Reasoning.}
We observe a decoupling between a model's proficiency in math reasoning and culturally situated tasks.
For instance, \textit{DS-R1-Distill-Qwen-7B} substantially outperforms \textit{DS-R1-Distill-Llama-8B} on \textit{MGSM} ($+19.69\%$) and \textit{MT-AIME} ($+11.39\%$), yet trails significantly in cultural reasoning ($-18.84\%$ on \textit{FORK}; $-18.56\%$ on \textit{CulturalBench}).
Besides, \textit{o4-mini-high} leads \textit{DeepSeek-V3.2} on \textit{MGSM} but exhibits worse performance in cultural reasoning.
This pattern suggests that the underlying reasoning mechanisms leveraged for structured, procedural calculation (Math) are distinct from those required for contextual, knowledge-intensive reasoning (Culture).

\paragraph{(2) \textbf{\textit{x1}} Models Exhibit Better Overall Reasoning Performance.}
Our results show that \textit{x1} systematically outperforms their backbone models in both multilingual and cultural scenarios. 
The improvements validate the efficacy of adaptive multilingual reasoning that leverages linguistic diversity in reasoning.
Notably, the gains are more pronounced in cultural tasks, suggesting a closer connection between cultural knowledge and linguistic expression of reasoning.
Moreover, despite utilizing only \textit{Think} mode data formats for self-generated training, \textit{x1} models show no noticeable performance degradation in the \textit{Non-Think} mode and even exhibit improvements in cultural scenarios.

\paragraph{(3) Greater Relative Improvements on Smaller and Weaker Backbones.}
We find that performance gains are inversely correlated with backbone strength. Smaller and lower-performing models, such as \textit{Qwen3-4B} and the \textit{DS-R1-Distill} series, achieve substantially larger relative improvements, whereas the gap between \textit{x1}-Qwen3-32B and its backbone is the smallest within the Qwen family. This trend suggests that as model scale and capability increase, multilingual and multi-domain reasoning becomes more internally integrated, leaving less "room" for improvement through thinking-language switching (We conduct a further discussion in Section~\ref{sec:net_benefits}).

\begin{figure*}[t]
    \centering
    \includegraphics[width=1\textwidth]{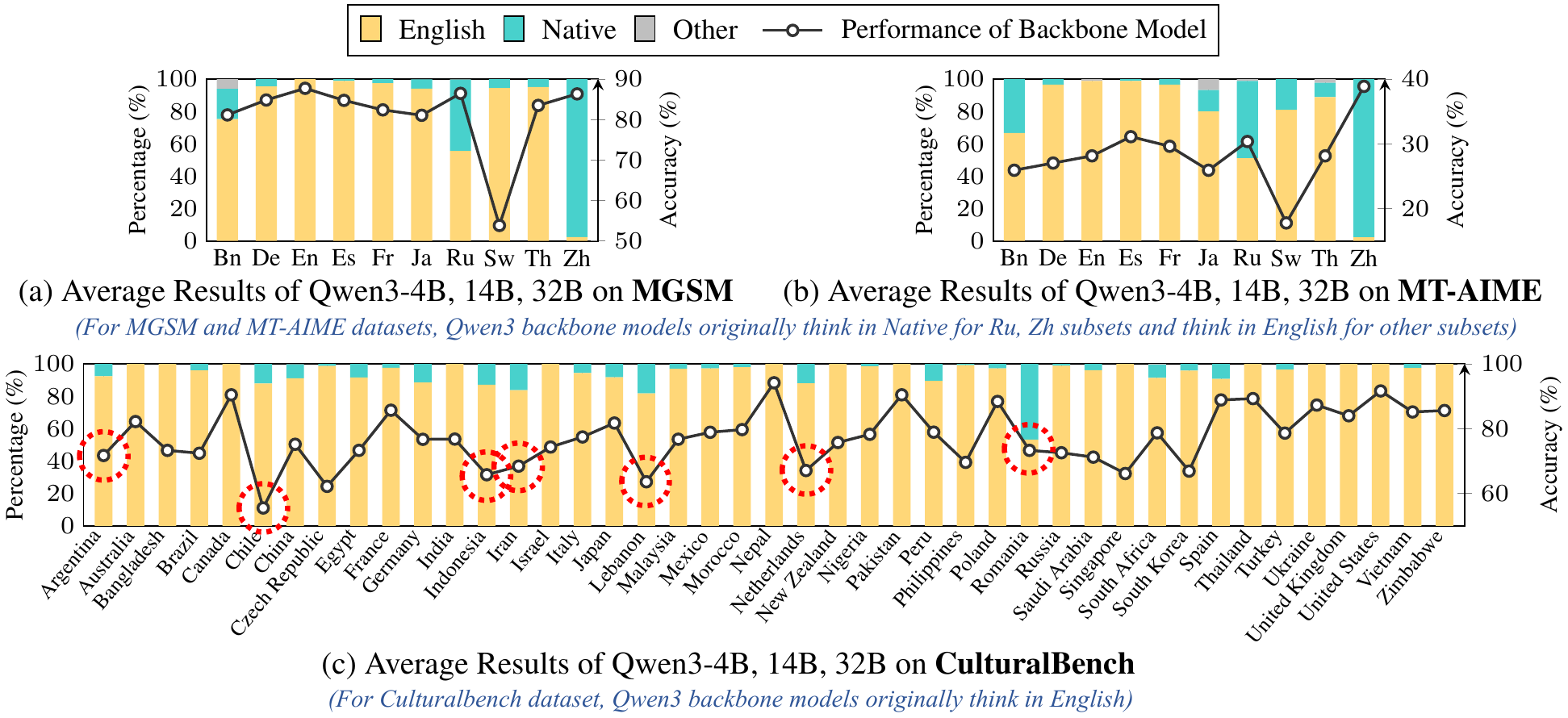}
    \caption{Distribution of thinking languages frequency by \textit{x1} models along with the performance of their backbones across different languages/cultures subsets. All statistics and results are averaged over \textit{Qwen3-4B}, \textit{14B} and \textit{32B}.
    }
    \label{fig:langdist}
\end{figure*}

\section{Further Analysis}\label{sec:analysis}

\subsection{Language Compliance in \textit{x1}}

We define \textit{Language Compliance} as the model's ability to satisfy the following requirements:
(1) \textit{Thinking Phase}: perform reasoning in a specified language.
(2) \textit{Answer Phase}: generate the final answer in the prompting language given the context, without being influenced by the reasoning language.
(3) \textit{Both Phases}: meet above two requirements simultaneously.
The language compliance results on \textit{MGSM} dataset are shown in Table~\ref{tab:lang_compliance}.

The results indicate that \textit{x1} models generally demonstrate great language compliance. 
However, a critical observation is the high correlation between the language compliance of \textit{x1} models and their backbone models. Since \textit{x1} models are trained on self-generated data from their backbones, pre-existing compliance deficiencies are not fully mitigated. This dependency is most pronounced in low-resource languages like Swahili, where the compliance rates for the Qwen3 backbones (4B: $26.4\%$, 14B: $49.6\%$, 32B: $64.0\%$) and \textit{x1} models (4B: $23.6\%$, 14B: $54.4\%$, 32B: $64.8\%$) remain low (We exclude the responses of \textit{x1-Qwen3-14B} on \textit{Bn} from subsequent analysis due to insufficient language compliance).

\subsection{Thinking-Language Switching in \textit{x1}}\label{sec:switching}

\subsubsection{Thinking-Language Frequency}

Figure~\ref{fig:langdist} illustrates the distribution of the chosen thinking language by \textit{x1} models and the vanilla performance of backbone models across various languages/cultures. We observe a key pattern: \textbf{\textit{x1} demonstrate an increased frequency of ``Native Thinking'' on language/culture subsets where backbone models originally perform worse}.

This pattern holds consistently across both multilingual math and cultural tasks, 
with the exception that Qwen3 backbone models naturally reason in Russian and Chinese for the corresponding math problems, leading \textit{x1} to retain a high proportion of native thinking for these languages.
For other languages/cultures, native-thinking frequency is inversely correlated with the backbone's original performance (e.g., in low-resource languages like Bengali, Swahili, Thai in Figure~\ref{fig:langdist} (a)(b), or low-performing cultural subsets circled in Figure~\ref{fig:langdist} (c)).
This trend suggests that when the dominant English reasoning pathway becomes unreliable, the model increases thinking-language switching as a recovery mechanism, attempting to exploit language-specific knowledge or alternative reasoning pathways encoded in non-English representations.

\subsubsection{Net Benefits of Language Switching}\label{sec:net_benefits}

\begin{figure*}[t]
    \centering
    \includegraphics[width=1\textwidth]{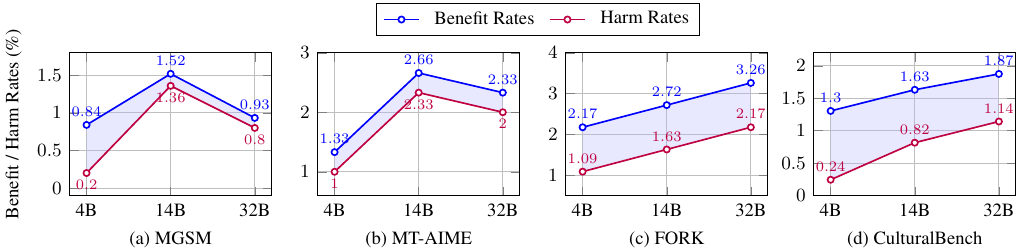}
    \caption{Benefit and harm rates of thinking-language switching on \textit{MGSM}, \textit{MT-AIME}, \textit{FORK} and \textit{CulturalBench} datasets across \textit{Qwen3-4B}, \textit{14B} and \textit{32B}. The shaded area (benefit rate minus harm rate) represents the net benefits.
    }
    \vspace{-0.3\baselineskip}
    \label{fig:hitcurves}
\end{figure*}

To assess the true utility of \textit{x1}'s adaptive thinking mechanism, we focus on the subset of samples where a thinking-language switch occurred. We define the \textit{Benefit Rate} as the proportion of cases where the backbone fails but \textit{x1} succeeds after switching (computed over all samples), and the \textit{Harm Rate} as the opposite case. Their difference reflects the \textit{net benefit} of adaptive switching. As shown in Figure~\ref{fig:hitcurves}, the benefit consistently outweighs the harm, providing clear evidence for the effectiveness of thinking-language switching. Besides, the analysis further reveals several insights:

\paragraph{(1) Larger Cross-Lingual Disparity Leads to Greater Net Benefits from Thinking-Language Switching.}

From Figure~\ref{fig:hitcurves} (a), we observe that in \textit{MGSM} (grade school level), the net benefit of thinking-language switching decreases steadily as model size scales. But for the more challenging benchmark \textit{MT-AIME} shown in (b), the net benefits remains low and nearly unchanged.

This trend aligns with the standard deviation (\textit{Std}) statistics in \S\ref{app:std} Table~\ref{tab:std}, where a larger \textit{Std} indicates greater cross-lingual performance disparity. We find a clear positive correlation: the greater this disparity, the larger the net benefit of switching.

For \textit{MGSM}, the Qwen3 backbone becomes increasingly balanced across languages as model size grows (Std: 4B-$16.96$, 14B-$7.446$, 32B-$4.665$), naturally reducing the benefit of switching. Intuitively, if performance were identical across all languages, switching would offer no gain. For \textit{MT-AIME}, although overall accuracy improves with scale, the performance imbalances remains low and stable (Std: 4B-$6.576$, 14B-$8.308$, 32B-$7.495$), explaining the consistently small and almost scale-invariant net benefit observed in Figure~\ref{fig:hitcurves}(b).

\paragraph{(2) Thinking in Culture-associated Language Offers Inherent and Persistent Advantages.}

Figure~\ref{fig:hitcurves} (c)(d) shows that the net benefit of reasoning in culture-associated language remains stable across model scales, and is consistently larger than that observed in multilingual math reasoning.
This indicates that the advantage of reasoning in culture-associated language is not a transient effect of limited model capacity, but instead reflects a deeper entanglement between language and culture.

\textbf{This persistence also reveals a fundamental challenge to ``Scaling Law''}: cultural knowledge is tightly intertwined with the linguistic forms through which it is acquired, represented, and invoked. Consequently, reasoning in English cannot fully substitute for reasoning in the culture-associated language, even as model size and overall capability continue to increase. This points to an intrinsic upper bound on English-centric reasoning in cultural contexts—one that is difficult to overcome through scaling alone.

\definecolor{ForestGreen}{RGB}{34,139,34}

\begin{table}[t]
    \small
    \renewcommand{\arraystretch}{1.2} 
      \centering
      \resizebox{\columnwidth}{!}{
        \begin{tabular}{rcc}
        \toprule
        {Models} & \makecell{Avg. Cultural Knowledge \\ Recall Count per Thought \\ (Efficiency)} & \makecell{Avg. Cultural Knowledge \\ Recall Accuracy (\%) \\ (Accuracy)} \\
        \midrule
        Qwen3-4B & 3.366 & 47.699 \\
        \textbf{\textit{x1}}-Qwen3-4B & 2.662 (\textcolor{red}{$\bm{\downarrow}$} 0.704) & 53.439 (\textcolor{ForestGreen}{$\bm{\uparrow}$} 5.740)      \\
        \cdashline{1-3}\noalign{\vskip 0.4ex}
        Qwen3-14B & 3.353 & 49.123 \\
        \textbf{\textit{x1}}-Qwen3-14B & 2.765 (\textcolor{red}{$\bm{\downarrow}$} 0.588) & 55.319 (\textcolor{ForestGreen}{$\bm{\uparrow}$} 6.196) \\
        \cdashline{1-3}\noalign{\vskip 0.4ex}Qwen3-32B & 3.561 & 45.813 \\
        \textbf{\textit{x1}}-Qwen3-32B & 3.000 (\textcolor{red}{$\bm{\downarrow}$} 0.561) & 56.725 (\textcolor{ForestGreen}{$\bm{\uparrow}$} 7.902) \\
      
        \bottomrule
        \end{tabular}
      }
    \caption{Statistics of cultural knowledge recall frequency and accuracy on the \textit{CulturalBench} dataset.
    }
    \vspace{-0.5\baselineskip}
    \label{tab:recall}
\end{table}

\subsection{Cultural Knowledge Recall Behavior}\label{sec:recall}

Section~\ref{sec:net_benefits} demonstrates the persistent advantages of thinking in culture-associated language. But what underlies this advantage?
As illustrated in Appendix \S\ref{app:recall} Figure~\ref{fig:recallcases}, both backbone and \textit{x1} models exhibit explicit \emph{cultural knowledge recall} during cultural reasoning, actively retrieving relevant cultural facts or norms as part of their reasoning process.
While such recall alone does not guarantee correctness, it plays a critical role in shaping the final answer.
Motivated by this observation, we compare the cultural knowledge recall behavior of \textit{x1} and its backbone models.
Specifically, we employ \textit{GPT-5-mini} to identify recalled cultural knowledge that supports or justifies the golden answer and to verify its correctness (implementations are in Appendix~\S\ref{app:recall}).
The results are summarized in Table~\ref{tab:recall}, leading to the following finding:

\paragraph{Thinking in Culture-associated Language Enables More Efficient and Accurate Cultural Knowledge Recall.}
As shown in Table~\ref{tab:recall}, across all model scales, \textit{x1} consistently recalls fewer but more accurate cultural knowledge instances compared to its backbone model. \textit{x1} reduces the average recall count per thought by $0.56$-$0.70$, while improving recall accuracy by $5.74\%$-$7.90\%$.
This indicates that reasoning in culture-associated languages promotes more targeted and relevant knowledge retrieval, rather than broader but noisier recall, providing a concrete mechanistic explanation for the persistent gains observed in cultural tasks.

\subsection{Language-Mixing in \textit{x1}}
In multilingual reasoning, models often interleave multiple languages within a single reasoning trajectory, a phenomenon commonly known as \textit{language mixing}~\citep{yong2025crosslingual} (we present a example in Figure~\ref{fig:example_langmix}).
Unlike explicit thinking-language switching, which operates at the context level, language mixing captures a finer-grained, sentence-level adaptation that can occur without explicit switching.
To isolate this effect, we focus on samples where \textit{x1} does not perform explicit thinking-language switching.
For these samples, we measure the change in sentence-level language-mixing strength of \textit{x1}'s responses relative to its backbone and analyze its correlation with the resulting net benefits aggregated over all samples.

\paragraph{Increased Language Mixing Is Associated with Positive Outcomes in Multilingual Reasoning.}

As we do not introduce any design about language mixing, we observe that on both \textit{MGSM} and \textit{MT-AIME}, the proportions of samples with increased and decreased language mixing rates are broadly balanced. However, those with increased language mixing rates consistently yield higher net benefits than those with decreased mixing ($0.26\%$ vs. $0.21\%$ on \textit{MGSM}, $0.85\%$ vs. $0.11\%$ on \textit{MT-AIME}).

These results indicate that the gains of \textit{x1} cannot be fully attributed to explicit language switching. Even in the absence of such switching, higher sentence-level language mixing rates is positively correlated with improved performance. 
Language mixing thus appears to act as a fine-grained mechanism that integrates complementary linguistic representations within a single reasoning trajectory, and may further provide a useful guidance signal for model optimization (e.g., in reward design).

\begin{figure}[t]
    \centering
    \includegraphics[width=0.4\textwidth]{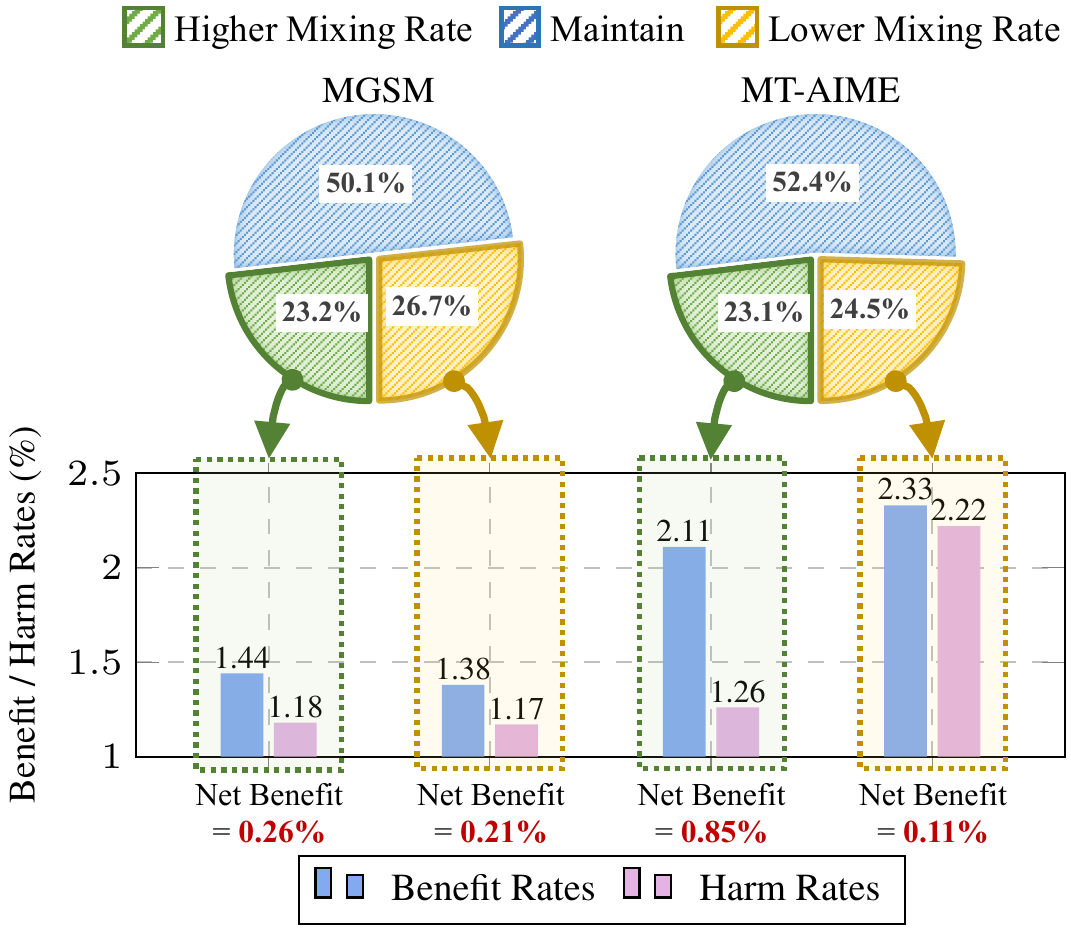}
    \caption{Proportions of samples with changed language mixing strength in \textit{x1}'s responses relative to its backbone, and the correlation with the resulting net benefits.}
    \vspace{-0.5\baselineskip}
    \label{fig:langmix}
\end{figure}

\subsection{Preference Optimization with DPO}

The instance-level contrastive signals naturally induce implicit preferences between alternative reasoning-language choices.
In this subsection, we further apply Direct Preference Optimization (DPO)~\citep{rafailov2023direct} as an auxiliary enhancement, treating reasoning trajectories in the advantageous language as preferred samples and those in the less advantageous language as dispreferred ones\footnote{Reusing the same data produced in Section~\ref{sec:method} Step~2.}.
The results are shown in Appendix \S\ref{app:dpo} Table~\ref{tab:main_dpo}, where we observe that DPO further improves the performance of \textit{x1} models, achieving a second jump in multilingual math reasoning.

\section{Related Work}\label{sec:related}

\paragraph{Multi- \& Cross-lingual Reasoning.}
Recent studies examined how LLMs perform reasoning across languages. 
Some approaches utilize response alignment, latent connection and language separability to bridge the gap between languages~\citep{zhu2024power,ye-etal-2025-cc,ye2025langgps}.
Besides, \citet{ko2025understand} anchor reasoning in English to minimize the gaps, whereas SLAM~\citep{fan2025slam} maintains reasoning capabilities with lower computational costs.
Further refinement is achieved through reasoning-focused tuning like mCoT~\citep{lai-nissim-2024-mcot}, preference-based alignment~\citep{she-etal-2024-mapo}, and process reward modeling~\citep{wang2025demystifying}.
Some previous works focus on prompting strategies~\citep{qin2023cross}.
Researchers proposed $\mathcal{X}$Transplant for cross-lingual complementarity~\citep{ye2024exploring}, dictionary insertion prompting~\citep{lu2024dictionary} that embeds key English terms into native prompts and program-based demonstrations~\citep{ranaldi2025natural} for structured logic transfer.
Recently, \citet{yong2025crosslingual} conducted a detailed analysis, revealing the potentials and limitations of cross-lingual test-time scaling.

\paragraph{Cultural Reasoning.}
Early efforts primarily focused on constructing culturally grounded knowledge bases, such as CANDLE~\citep{nguyen2023extracting} for normative assertions, CultureBank~\citep{shi2024culturebank} for community-driven cultural descriptors, MAPS~\citep{liu2024multilingual} for figurative expressions, and GlobeSumm~\citep{ye-etal-2024-globesumm} for global perspectives on international news.
From a methodological perspective, existing work explores how to elicit or control cultural perspectives in LLMs~\citep{kovavc2023large}, enhance cultural reasoning by explicitly recognizing cultural context in prompts~\citep{wang-etal-2024-countries}, integrate moral reasoning across normative ethics frameworks~\citep{rao-etal-2023-ethical}, and complement English reasoning with non-English cultural features via cross-lingual transplantation~\citep{ye2024exploring}.

Rather than enforcing consistency across languages, we investigate whether different languages inherently provide complementary reasoning advantages for different instances.
By framing reasoning-language choice as a self-aware decision process, our method bridges multilingual and cultural reasoning from a unified perspective, positioning language as an active component of reasoning rather than a passive carrier of content.

\section{Conclusion}

In this work, we presented \textit{x1} models that enable adaptive multilingual reasoning.
By constructing \textit{x1} without expanding the model's knowledge boundaries, we isolated the effect of reasoning-language choice and demonstrated that linguistic diversity can be systematically leveraged to improve both multilingual mathematical reasoning and culturally grounded reasoning.
Beyond empirical gains, our findings challenge a simplistic interpretation of scaling laws, revealing that even as models grow larger and more capable, the advantages of reasoning in culture-associated languages persist rather than vanish.
This highlights reasoning-language choice as a functional component of reasoning rather than a superficial artifact of generation.
We hope this work encourages future research to move beyond monolithic reasoning paradigms toward building LLMs that are more globally and culturally competent.

\section*{Limitations}
This work exhibits several limitations worth noting.
First, to enable a clean comparison between alternative reasoning pathways, for each instance, we intentionally restrict the candidate thinking languages to English and one non-English language (which is determined by the query language or cultural background).
Although this design choice facilitates controlled analysis, it limits the exploration of richer multilingual combinations and linguistic diversity.
Extending adaptive reasoning to a larger pool of languages can be a further exploration.
Second, while our analysis reveals the potential positive impact of ``language mixing'' for reasoning, we do not pursue this direction further due to space constraints.
Exploring principled ways to incorporate mixed-language reasoning, for example as a training signal or an optimization objective can be great future work.
Thirdly, for culture-related tasks, we employ an LLM-as-a-Judge framework to assess response quality.
Although this approach enables scalable evaluation, it may inherit the biases and limitations of the evaluator model.
We mitigate this risk by using evaluation only for relative comparison between reasoning-language choices, rather than for absolute or normative judgments.

\section*{Acknowledgements}
Xiaocheng Feng is the corresponding author of this work.
We thank the anonymous reviewers for their insightful comments.
This work was supported by the National Natural Science Foundation of China (NSFC) (grant 62522603, 62276078), the Key R\&D Program of Heilongjiang via grant 2022ZX01A32, the Fundamental Research Funds for the Central Universities ( XNJKKGYDJ2024013 ) .
\section*{Ethical Considerations}
This work studies multilingual and culturally grounded reasoning in large language models.
All model outputs, especially those related to cultural knowledge or perspectives, are generated by the models themselves and do not represent the views or opinions of the authors.
Our method does not introduce new knowledge sources or external supervision, and therefore does not increase the risk of generating sensitive or private information beyond the capabilities of the backbone models.
While reasoning in culture-associated languages may surface different cultural expressions or viewpoints, such outputs should not be interpreted as authoritative or normative cultural judgments.
In this paper, we use Gemini to correct grammatical errors.

\bibliography{custom}

\clearpage
\appendix

\section{Experiment Details}\label{app:exp}

\subsection{Step 1: To be a Multilingual Reasoner}\label{app:step1}

\paragraph{Language List.}
The 31 languages (English included) involved in step~1 training are as follows:
\begin{lstlisting}[language=Python, basicstyle=\scriptsize, belowskip=1pt]
["Arabic","Bulgarian","Bengali","German","Greek",
"English","Spanish","Finnish","French","Hebrew",
"Hindi","Hungarian","Indonesian","Italian",
"Japanese","Korean","Malay","Dutch","Polish",
"Portuguese","Romanian","Russian","Swedish",
"Swahili","Thai","Tagalog","Turkish","Ukrainian",
"Urdu","Vietnamese","Chinese"]
\end{lstlisting}
\paragraph{Quality Filtering.}
To ensure linguistic fidelity, we apply quality filtering using COMET (Unbabel/wmt23-cometkiwi-da-xxl) scores and discard translations of evidently poor quality (whose scores fall below 0.4).

\begin{figure*}[t]
    \centering
    \includegraphics[width=1\textwidth]{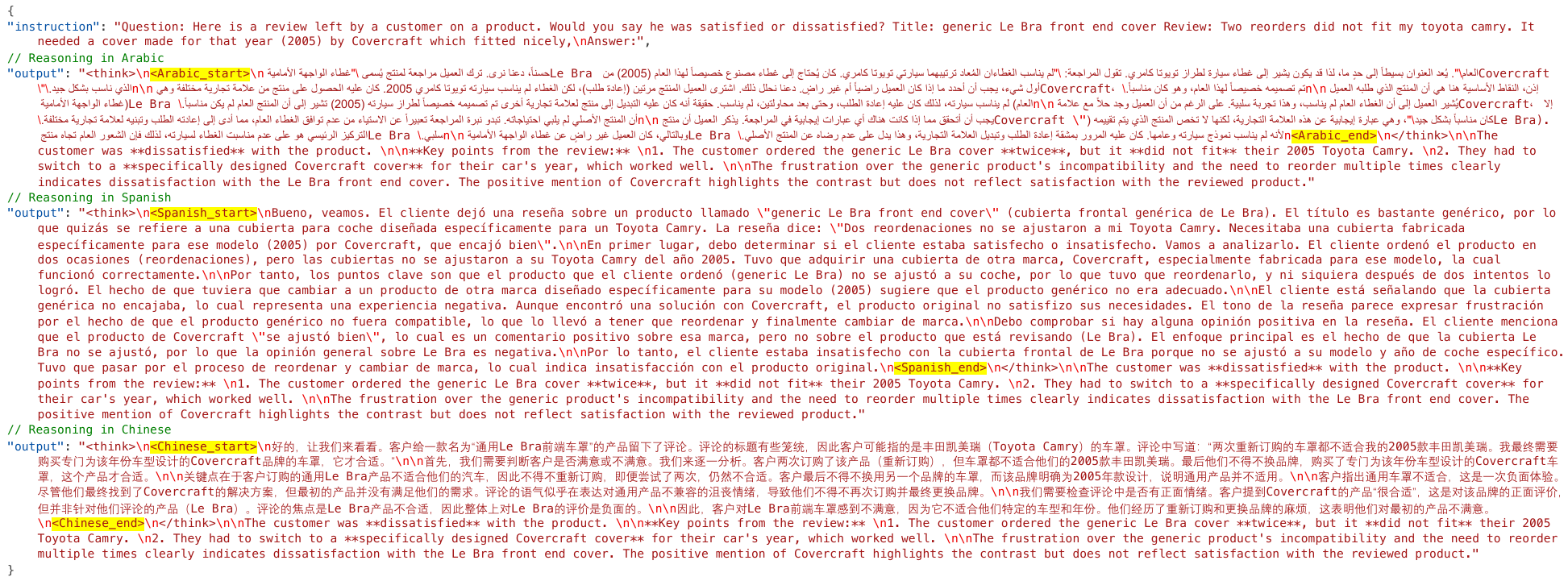}
    \caption{Examples of Step 1 constructed training data (reasoning in Arabic, Spanish and Chinese).
    }
    \label{fig:step1_cases}
\end{figure*}

\paragraph{Data Format.}
We adopt the official training templates provided by the corresponding backbone models.
For instance, when training \textit{Qwen3-4B}, \textit{14B}, and \textit{32B}, we use the official \textit{``Qwen3''} template to ensure consistency with the original training and inference protocols.
Beyond the official templates, we introduce a customized template tailored for multilingual reasoning control.
This template explicitly specifies the language used in the reasoning process while keeping the final answer aligned with the prompt language.
Real examples are shown in Figure~\ref{fig:step1_cases}.
Note that, due to the diversity of $X$ and to preserve the generality of the template, we do not register ``\textit{<$X$\_start>}'' series as special tokens in the tokenizer.

\paragraph{Training Hyperparameters.}
We perform full-parameter finetuning for all models, except for \textit{Qwen3-32B}, for which we adopt LoRA training due to its large model size.
\begin{lstlisting}[language=Python, basicstyle=\scriptsize, belowskip=1pt]
# for all models except Qwen3-32B
per_device_train_batch_size: 1
gradient_accumulation_steps: 16
learning_rate: 1.0e-5
num_train_epochs: 3.0
bf16: true
\end{lstlisting}

\begin{lstlisting}[language=Python, basicstyle=\scriptsize, belowskip=1pt]
# for Qwen3-32B
lora_rank: 8 # default lora_alpha: lora_rank * 2
lora_target: all
per_device_train_batch_size: 1
gradient_accumulation_steps: 16
learning_rate: 1.0e-4
num_train_epochs: 3.0
bf16: true
\end{lstlisting}

\subsection{Step 2: To be an Adaptive Reasoner}\label{app:step2}
\paragraph{Training Data Statistics.}
\begin{itemize}[leftmargin=*]
\setlength{\parsep}{0pt}
\setlength{\parskip}{0pt}
\item (1) Multilingual Math Problems: We use the \textit{MGSM8KInstruct} dataset~\citep{chen2023breaking}, which contains math problems paired with their correct answers. We sample 200 questions across 10 languages, resulting in 2,000 training instances.
Detailed statistics are as follows:
\begin{lstlisting}[language=Python, basicstyle=\scriptsize, belowskip=0pt]
{   // language: sample size
    "Bengali": 200,
    "German": 200,
    "English": 200,
    "Spanish": 200,
    "French": 200,
    "Japanese": 200,
    "Russian": 200,
    "Swahili": 200,
    "Thai": 200,
    "Chinese": 200
}
// Total: 2000
\end{lstlisting}
\item (2) Culture-related Problems: We use the \textit{CultureBank} dataset~\citep{shi2024culturebank}, which provides cultural questions along with the underlying cultural knowledge they reflect. We sample 100-200 questions from each of the 25 language groups (depending on availability), covering cultural questions from 45 countries/regions, resulting in 4,413 samples in total.
Detailed statistics are as follows:
\begin{lstlisting}[language=Python, basicstyle=\scriptsize, belowskip=0pt]
{   // language group: sample size
    "Arabic": 200,
    "Danish": 122
    "German": 200,
    "Greek": 100,
    "English": 200,
    "Spanish": 200,
    "Finnish": 142,
    "French": 200,
    "Irish": 157,
    "Scottish Gaelic": 125,
    "Hindi": 200,
    "Indonesian": 200,
    "Italian": 200,
    "Japanese": 200,
    "Korean": 200,
    "Maori": 141,
    "Malay": 130,
    "Dutch": 200,
    "Norwegian": 173,
    "Polish": 123,
    "Portuguese": 200,
    "Russian": 200,
    "Swedish": 200,
    "Tagalog": 200,
    "Chinese": 200
}
// Total: 4413
\end{lstlisting}
Each language group is associated with a list of countries/regions:
\begin{lstlisting}[language=Python, basicstyle=\scriptsize, belowskip=-1pt]
{   // language group: list of countries/regions
    "Arabic":["Algeria","Arab","Egypt","Iraq",
    "Lebanon","Syria","Morocco","Tunisia","Jordan"],
    "Danish":["Denmark"],
    "German":["Germany","Austria","Switzerland"],
    "Greek":["Greece","Cyprus"],
    "English":["United States"],
    "Spanish":["Argentina","Spain","Cuba",
    "Chile","Colombia","Dominican Republic",
    "Mexico","Peru"],
    "Finnish":["Finland"],
    "French":["France"],
    "Irish":["Ireland"],
    "Scottish Gaelic":["Scotland"],
    "Hindi":["India"],
    "Indonesian":["Indonesia"],
    "Italian":["Italy"],
    "Japanese":["Japan"],
    "Korean":["Korea"],
    "Maori":["New Zealand"],
    "Malay":["Malaysia"],
    "Dutch":["Netherlands","Belgium"]
    "Norwegian":["Norway"],
    "Polish":["Poland"],
    "Portuguese":["Brazil","Portugal"],
    "Russian":["Russia"],
    "Swedish":["Sweden"],
    "Tagalog":["Philippines"],
    "Chinese":["China"]
}
\end{lstlisting}
\end{itemize}

\paragraph{Advantageous Language Identification.}
For each question in $D_{\operatorname{math}}$ or $D_{\operatorname{culture}}$, we first obtain the paired reasoning trajectories in both English and non-English with $\mathcal{M}$ and $\mathcal{M}_{\operatorname{surface}}$\footref{myfoot}.
We identify the advantageous reasoning language based on the quality of the final answers, under the principle that stronger reasoning should manifest in superior outputs.
For multilingual math problems, answer quality is determined via numerical matching.
\begin{equation}
    score_{\text{math}}(\hat{y}, y) =
    \begin{cases}
    10, & \text{if } \hat{y} \text { matches } y, \\
    0,  & \text{otherwise},
    \end{cases}
\end{equation}
where $\hat{y}$ denotes the model-predicted numerical answer and $y$ is the ground-truth answer.

For culture-related questions, we employ \textit{GPT-5-mini} to score how well the response entails the relevant cultural knowledge.
\begin{equation}
\begin{aligned}
&score_{\text{culture}}(\hat{y}, n) = \text{Entailment}(Q, \hat{y}, n, LLM) \\
&\hspace{2.6cm} \in [0, 10]
\end{aligned}
\end{equation}
where $Q$ denotes the culture-related question, $\hat{y}$ denotes the model-predicted response and $n$ represents the relevant cultural knowledge or norm that the model is expected to comply with.
The prompt for scoring is shown in Table~\ref{tab:prompt_score}.

\definecolor{color1}{RGB}{240,230,140}
\definecolor{color2}{RGB}{197,217,197}
\definecolor{color3}{RGB}{225,179,191}
\definecolor{color4}{RGB}{176,224,230}

\begin{table}[t]
    \renewcommand{\arraystretch}{0.9} 
    \small
    \centering 
    \setlength\tabcolsep{3pt}
    \resizebox{0.9\columnwidth}{!}{%
        \begin{tabular}{p{\columnwidth}}
        \toprule
        Prompt for Scoring Cultural Knowledge Entailment \\
        \toprule
        You will be given a culture-related question, a related cultural knowledge and a response from an agent.\\
        \\
        Your task is to provide a 'total rating' scoring how well the response from an agent answers the culture-related question and entails the related cultural knowledge.\\
        \\
        Give your answer as a float on a scale of 0 to 10, where 0 means that the response from agent is not helpful at all, and 10 means that the answer completely and helpfully addresses the question.\\
        \\
        Your output should be a float score only.\\
        \\
        Question: \colorbox{color2}{\{question\}} \\
        \\
        Cultural Knowledge: \colorbox{color2}{\{cultural knowledge description\}}  \\
        \\
        Response: \colorbox{color2}{\{model response\}} \\
        \\
        Total rating (a float on a scale of 0 to 10): \\
        \bottomrule
        \end{tabular}
    }
    \caption{Prompts designed for scoring cultural knowledge entailment.}
    \label{tab:prompt_score}
\end{table}

The reasoning language associated with the higher-scoring trajectory is then regarded as the advantageous thinking language for that instance.

\begin{figure}[h]
    \centering
    \includegraphics[width=0.42\textwidth]{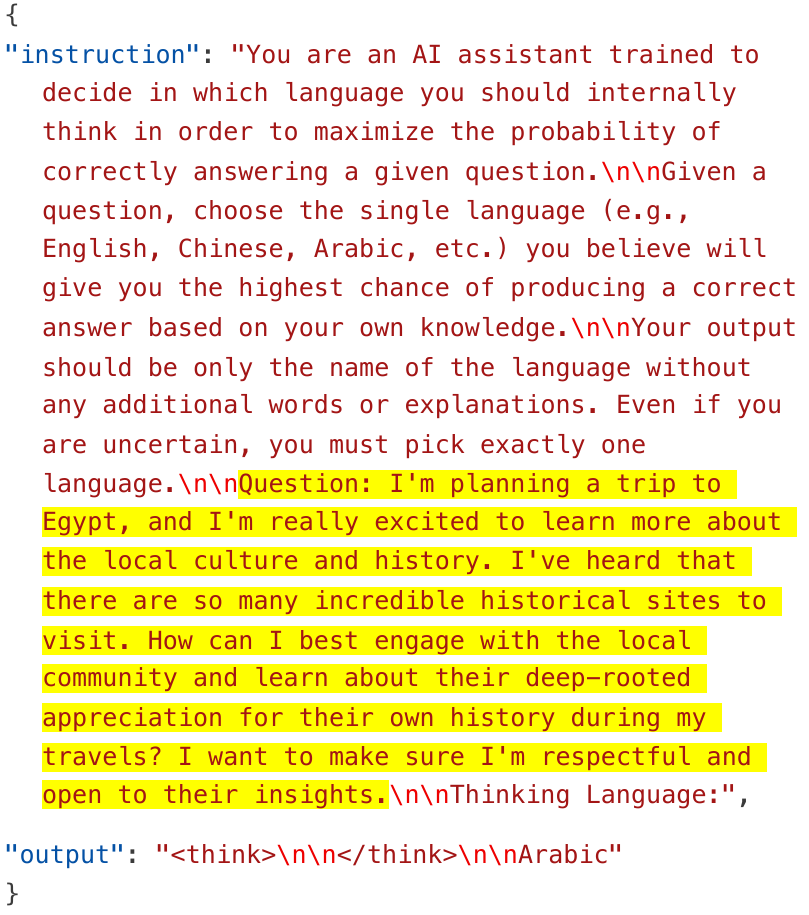}
    \caption{Example of self-awareness data, where ``Arabic'' is the advantageous reasoning language.
    }
    \label{fig:step2_cases}
\end{figure}

\paragraph{Data Format.}
Besides the data format introduced in Figure~\ref{fig:step1_cases}, we also construct self-awareness data that externalizes the model's language-selection decision as an explicit prediction task and the format is shown in Figure~\ref{fig:step2_cases}.

\paragraph{Training Hyperparameters.}
We apply LoRA training separately for the multilingual math and cultural reasoning scenarios. For \textit{MGSM8KInstruct}, we set the LoRA rank to 4, as we empirically observe that using a larger rank leads to unexpected and substantial drops in performance.
\begin{lstlisting}[language=Python, basicstyle=\scriptsize, belowskip=-2pt]
# for MGSM8KInstruct
lora_rank: 4 # default lora_alpha: lora_rank * 2
lora_target: all
per_device_train_batch_size: 1
gradient_accumulation_steps: 16
learning_rate: 1.0e-4
bf16: true
\end{lstlisting}

\begin{lstlisting}[language=Python, basicstyle=\scriptsize, belowskip=-2pt]
# for CultureBank
lora_rank: 16 # default lora_alpha: lora_rank * 2
lora_target: all
per_device_train_batch_size: 1
gradient_accumulation_steps: 16
learning_rate: 1.0e-4
bf16: true
\end{lstlisting}

\subsection{Repetition-Detection Truncation Mechanism}\label{app:truncation}

To prevent the model from producing meaningless repetitive reasoning during inference, we implement a repetition-detection truncation mechanism based on a custom \texttt{StoppingCriteria}. The method monitors the generated text in fixed-size blocks and terminates decoding if a newly generated block has appeared earlier in the output, indicating repetitive looping behavior.

\begin{algorithm}[H]
\captionsetup{font=small, labelfont=small}
\small
\caption{Repetition-Detection Truncation Mechanism}
\begin{algorithmic}[1]
\raggedright
\Require Tokenizer $\mathcal{T}$, prompt length $L$, block size $B$, model $\mathcal{M}$
\State Initialize generated sequence $X \leftarrow$ model prompt
\State flag $\leftarrow$ False

\While{not flag}
    \State $x \leftarrow \mathcal{M}$\texttt{.generate\_next}( )
    \State Append $x$ to $X$
    \State $T \leftarrow \mathcal{T}.\texttt{decode}(X[L:])$ \Comment{decode newly generated text}
    \If{$|T| < 2B$}
        \State \textbf{continue} \Comment{insufficient text length}
    \EndIf
    \State last\_block $\leftarrow T[|T|-B : |T|]$
    \State prefix $\leftarrow T[0 : |T|-B]$
    \If{last\_block $\in$ prefix}
        \State flag $\leftarrow$ True \Comment{repetition detected}
        \State \textbf{break}
    \EndIf
\EndWhile

\State \Return $X$
\end{algorithmic}
\end{algorithm}

\definecolor{up-green}{RGB}{0,120,0}
\definecolor{down-red}{RGB}{255,0,0}
\definecolor{ForestGreen}{RGB}{34,139,34}

\begin{table}[t]
\small
\renewcommand{\arraystretch}{1.1} 
\setlength{\dashlinedash}{4pt} 
\setlength{\dashlinegap}{2pt}  
  \centering
  \resizebox{0.85\linewidth}{!}{
    \begin{tabular}{l@{\hskip 4pt}cccc}
    \toprule
    \multirow{3}[4]{*}{\textbf{Models}} & \multicolumn{4}{c}{\textbf{Multilingual Math Reasoning}} \\
    \cmidrule(lr){2-5}
    & \multicolumn{2}{c}{\textbf{MGSM}} & \multicolumn{2}{c}{\textbf{MT-AIME}} \\
    & \multicolumn{1}{c}{\textit{Non-Think}} & \multicolumn{1}{c}{\underline{\textbf{\textit{Think}}}} & \multicolumn{1}{c}{\textit{Non-Think}} & \multicolumn{1}{c}{\underline{\textbf{\textit{Think}}}}  \\
    \midrule
    \multicolumn{5}{c}{\textbf{\textit{Backbone Models}}} \\
    \midrule
    \text{Qwen3-4B} & 70.21 & 76.59 & 12.89 & 21.78 \\
    \text{Qwen3-14B} & 77.64 & 82.56 & 19.33 & 29.22  \\
    \text{Qwen3-32B} & 80.52 & 83.98 & 21.83 & 33.89  \\
    \midrule
    \multicolumn{5}{c}{\textbf{\textit{+ Vanilla Finetuning}}} \\
    \midrule
    {Qwen3-4B}
    & \multicolumn{2}{c}{50.80} & \multicolumn{2}{c}{1.667} \\
    {Qwen3-14B}
    & \multicolumn{2}{c}{64.24} & \multicolumn{2}{c}{2.000} \\
    {Qwen3-32B}
    & \multicolumn{2}{c}{72.68} & \multicolumn{2}{c}{0.333} \\
    \midrule
    \multicolumn{5}{c}{\textbf{\textit{Backbone + Majority Voting}}} \\
    \midrule
    {Qwen3-4B}
    & 87.46 & 92.00 & 16.67 & 35.56  \\
    {Qwen3-14B}
    & 89.33 & 92.93 & 28.89 & 47.78 \\
    {Qwen3-32B}
    & 89.20 & 93.07 & 34.44 & 50.00 \\
    \midrule
    \multicolumn{5}{c}{\textbf{\textit{x1 Series Models}}} \\
    \midrule
    {\textbf{\textit{x1}}-Qwen3-4B}
    & 70.30 & 77.69 & 13.56 & 22.83 \\
    {\textbf{\textit{x1}}-Qwen3-14B}
    & 77.38 & 83.64 & 19.44 & 33.11 \\
    {\textbf{\textit{x1}}-Qwen3-32B}
    & 80.12 & 84.43 & 22.11 & 34.50 \\
    \midrule
    \multicolumn{5}{c}{\textbf{\textit{x1 Series Models + Majority Voting}}} \\
    \midrule
    {\textbf{\textit{x1}}-Qwen3-4B}
    & 88.80 & 93.33 & 20.00 & 35.56 \\
    {\textbf{\textit{x1}}-Qwen3-14B}
    & 90.27 & 93.73 & 31.11 & 48.89 \\
    {\textbf{\textit{x1}}-Qwen3-32B}
    & 90.00 & 93.20 & 36.67 & 50.00 \\
    \bottomrule
    \end{tabular}
  }
  \caption{Supplementary experimental Mean@3 results that are the averages of the performance across all languages / cultures involved for each dataset.}
  \label{tab:main_supp}
\end{table}

\begin{table*}[t]
    \small
    \renewcommand{\arraystretch}{1.1} 
    \setlength{\tabcolsep}{4pt}
    \setlength{\dashlinedash}{4pt} 
    \setlength{\dashlinegap}{2pt}  
      \centering
      \resizebox{0.8\textwidth}{!}{
        \begin{tabular}{lcccccccccc|c|c}
            \toprule
            \multirow{3}{*}{Models} & \multicolumn{12}{c}{Mean@3 Performance on MGSM} \\
            \cmidrule(lr){2-13}
            & Bn & De & En & Es & Fr & Ja & Ru & Sw & Th & Zh & Average & \textbf{Standard Deviation} \\
            \midrule
            \text{Qwen3-4B} &  75.33 & 82.53 & 87.60 & 84.67 & 81.6 & 78.67 & 84.67 & 26.80 & 79.33 & 84.67 & 76.59 & \textbf{16.96} \\
            \text{Qwen3-14B} &  83.60 & 85.60 & 86.60 & 84.60 & 82.80 & 82.20 & 87.00 & 60.80 & 86.00 & 86.40 & 82.56 & \textbf{7.446} \\
            \text{Qwen3-32B} &  84.00 & 86.60 & 89.20 & 84.20 & 83.20 & 82.80 & 86.80 & 71.80 & 84.20 & 87.00 & 83.98 & \textbf{4.665} \\

            \toprule
            \multirow{3}{*}{Models} & \multicolumn{12}{c}{Mean@3 Performance on MT-AIME} \\
            \cmidrule(lr){2-13}
            & Bn & De & En & Es & Fr & Ja & Ru & Sw & Th & Zh & Average & \textbf{Standard Deviation} \\
            \midrule
            \text{Qwen3-4B} &  22.22 & 25.56 & 25.56 & 24.44 & 23.33 & 22.22 & 17.78 & 6.67 & 25.56 & 24.44 & 21.78 & \textbf{6.576} \\
            \text{Qwen3-14B} &  26.67 & 26.67 & 32.22 & 25.56 & 34.44 & 24.44 & 34.44 & 18.89 & 23.33 & 45.56 & 29.22 & \textbf{8.308} \\
            \text{Qwen3-32B} &  28.89 & 28.89 & 26.67 & 43.33 & 31.11 & 31.11 & 38.89 & 27.78 & 35.56 & 46.67 & 33.89 & \textbf{7.495} \\
            \bottomrule
        \end{tabular}

      }
      \caption{Performance of backbone models on \textit{MGSM }and \textit{MT-AIME} along with the average and standard deviation. The larger the standard deviation, the greater the cross-lingual performance disparity.}
      \label{tab:std}
    \end{table*}

\subsection{Results For Each Subset}\label{app:subset_exp}
We present the results for each language subset for \textit{MGSM} and \textit{MT-AIME} in Table~\ref{tab:main_math}.

\subsection{Supplementary Experimental Results}\label{app:supp_exp}
We present the supplementary experimental results of (1) \textit{vanilla finetuning} and (2) \textit{majority voting} in Table~\ref{tab:main_supp}.
Since the \textit{CultureBank} dataset lacks reference answers for supervised training, and \textit{FORK} and \textit{CulturalBench} are not designed to support multilingual evaluation, we restrict the analysis to multilingual mathematical reasoning tasks.

\begin{itemize}[leftmargin=*]
\setlength{\parsep}{0pt}
\setlength{\parskip}{0pt}
\item \textbf{Vanilla Finetuning:} We perform vanilla finetuning (lora) for the backbone models with \textit{MGSM8KInstruct} dataset.
\item \textbf{Majority Voting:} For each math problem, we query the model using 10 parallel versions of the same question in different languages and aggregate the resulting answers via majority voting, taking the most frequent answer as the final prediction.
In this setting, we ignore constraints on response-language consistency and the voting outcome serves as an approximate indicator of the potential gains from linguistic diversity.
\end{itemize}

We observe that continuous finetuning on already strong \textit{Qwen3}-series models leads to a noticeable degradation in performance.
This degradation is likely due to the limited difficulty and suboptimal quality of the reasoning traces in \textit{MGSM8KInstruct}, which can substantially disrupt the models' existing reasoning capabilities.

By contrasting ``\textit{Backbone + Majority Voting}'' with ``\textit{x1 + Majority Voting}'', we find that the upper bound afforded by linguistic diversity remains remarkably high: aggregating predictions across 10 languages substantially surpasses the models' original performance.
Moreover, the \textit{x1} series consistently exhibits a higher majority-voting upper bound than their backbone models, suggesting that \textit{x1} more effectively integrates and aligns reasoning signals across languages, thereby enabling linguistic diversity to be exploited more coherently.

\begin{figure*}[t]
    \centering
    \includegraphics[width=1\textwidth]{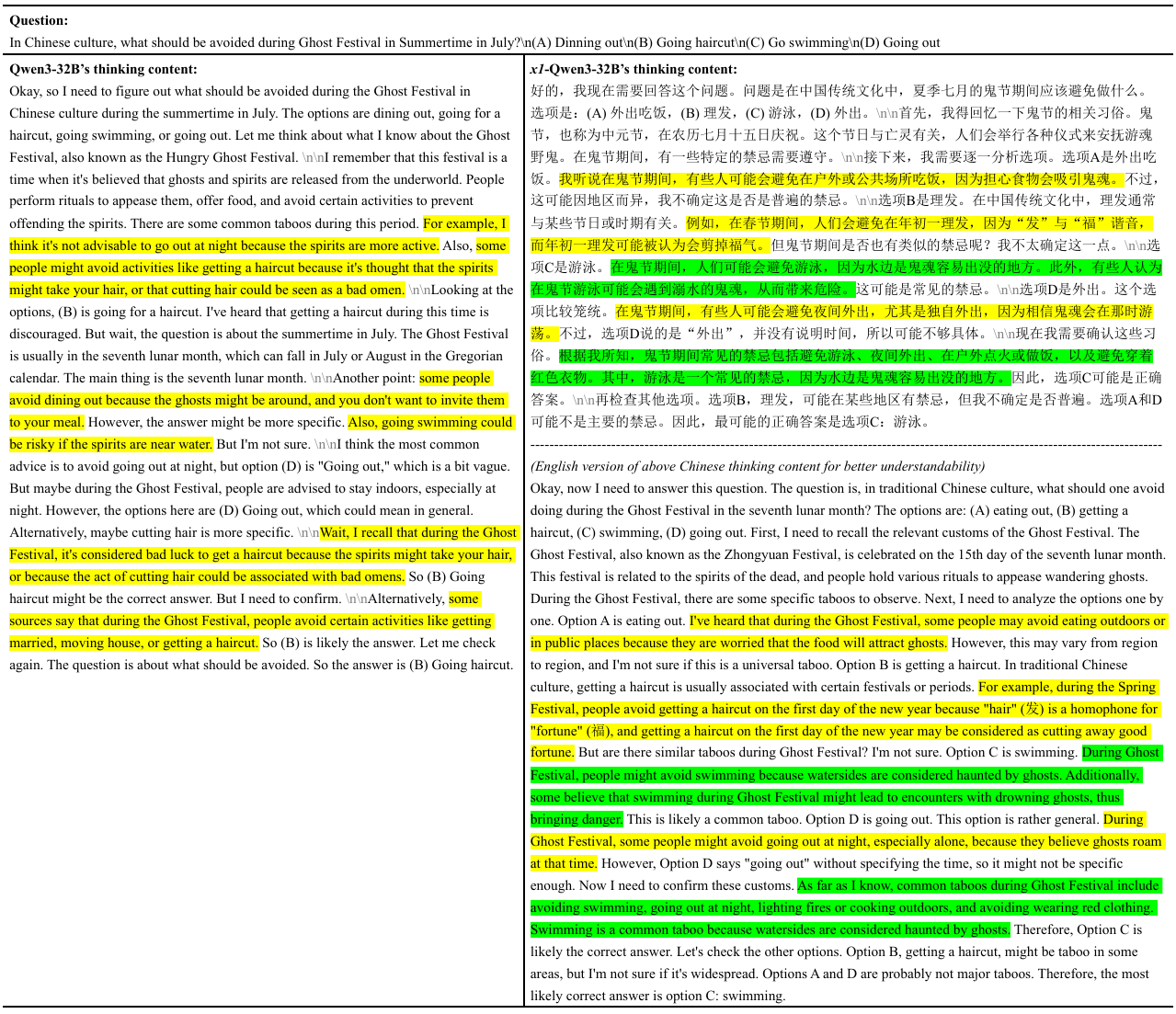}
    \caption{Examples of cultural knowledge recall behavior.
    }
    \label{fig:recallcases}
\end{figure*}

\definecolor{color1}{RGB}{240,230,140}
\definecolor{color2}{RGB}{197,217,197}
\definecolor{color3}{RGB}{225,179,191}
\definecolor{color4}{RGB}{176,224,230}

\begin{table*}[t]
    \small
    \centering 
    \setlength\tabcolsep{3pt}
    \setlength{\dashlinedash}{4pt} 
    \setlength{\dashlinegap}{2pt}  
    \resizebox{\textwidth}{!}{%
        \begin{tabular}{p{\textwidth}}
        \toprule
        Prompt for Identifying Cultural Knowledge Recall Behaviors \\
        \toprule
        (Input:) \\
        Given the following QA pair, I will provide you with my reasoning process. \\
        Your task is to determine whether my reasoning includes any recall of cultural norms that directly support or justify the Golden Answer. \\
        If such culturally relevant recall exists, extract only the portions that are directly tied to the Golden Answer and return them in a Python list.\\
        \\
        Question: \colorbox{color2}{\{question\}} \\
        \\
        Golden Answer: \colorbox{color2}{\{answer\}} \\
        \\
        Reasoning Process: \colorbox{color2}{\{reasoning\}} \\
        \noalign{\vskip 1ex}
        \cdashline{1-1}
        \noalign{\vskip 1ex}
        (Output:) \\
        \colorbox{color4}{\{Response of Extracted Cultural Knowledge Recalls\}} \\ 
        \bottomrule
        \noalign{\vskip 1ex}
        Prompt for Verifying Cultural Knowledge Recall Behaviors \\
        \toprule
        (Input:) \\
        Given the following QA pair, I will provide you with a cultural statement. \\
        Your task is to determine whether this statement provides a decisive and indispensable contribution to arriving at the Golden Answer. \\
        Mere relevance or weak association does not count. \\
        Return only True or False. \\
        \\
        Question: \colorbox{color2}{\{question\}}  \\
        \\
        Golden Answer: \colorbox{color2}{\{answer\}} \\
        \\
        Cultural Statement: \colorbox{color2}{\{norm\}} \\
        \noalign{\vskip 1ex}
        \cdashline{1-1}
        \noalign{\vskip 1ex}
        (Output:) \\
        \colorbox{color4}{\{True or False\}} \\ 
        \bottomrule
        \end{tabular}
    }
    \caption{Prompts for identifying and verifying cultural knowledge recall behaviors.}
    \label{tab:prompt_recall}
\end{table*}

\begin{figure*}[t]
    \centering
    \includegraphics[width=\textwidth]{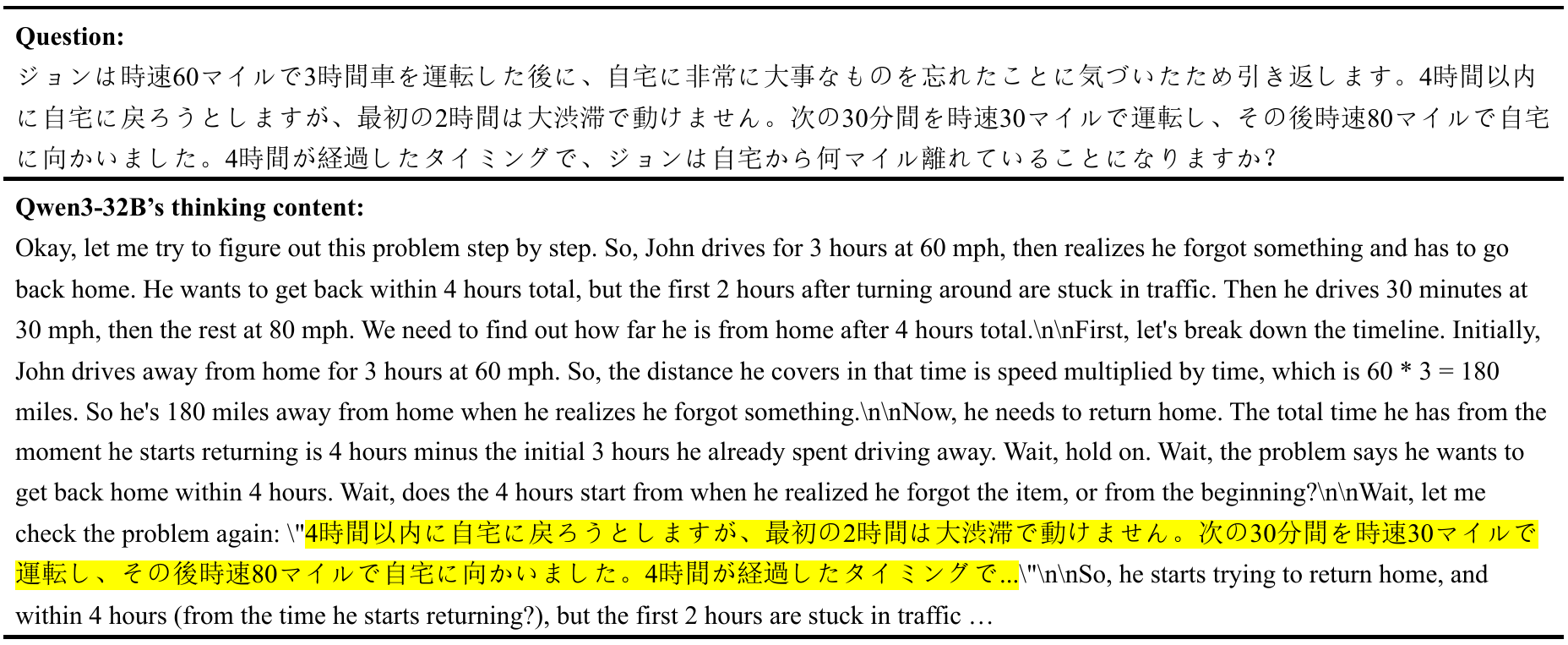}
    \caption{Example of language mixing behavior.
    }
    \label{fig:example_langmix}
\end{figure*}

\section{Performance Standard Deviation Across Languages}\label{app:std}

The statistics of performance standard deviation (\textit{Std}) across languages for backbone models in \textit{MGSM} and \textit{MT-AIME} is presented in Table~\ref{tab:std}. Larger \textit{Std} indicates greater cross-lingual performance disparity. The statistics provides important support for the analysis in Section~\ref{sec:net_benefits} (1).

\section{Cultural Knowledge Recall Analysis}\label{app:recall}

We observe that both backbone and \textit{x1} models exhibit explicit \emph{cultural knowledge recall} behavior during cultural reasoning, actively retrieving relevant cultural facts or norms as part of their reasoning process.
While such recall alone does not guarantee correctness, it plays a critical role in shaping the final answer.
As illustrated in Figure~\ref{fig:recallcases}, we present a pair of examples of cultural knowledge recall behavior produced by \textit{Qwen3-32B} (thinking in English) and \textit{x1-Qwen3-32B} (thinking in Chinese) on a question from \textit{CulturalBench} dataset.
We use high-light to highlight the cultural knowledge recall behavior, where the green parts are the key recall behaviors that directly lead to the model's correct answer.

Motivated by this observation and to further analyze such behavior, we employ \textit{GPT-5-mini} to identify recalled cultural knowledge that supports or justifies the golden answer and to verify its correctness. The designed prompts are presented in Table~\ref{tab:prompt_recall}.

As shown in Table~\ref{tab:recall}, across all model scales, \textit{x1} consistently recalls fewer but more accurate cultural knowledge instances compared to its backbone model. \textit{x1} reduces the average recall count per thought by $0.56$-$0.70$, while improving recall accuracy by $5.74\%$-$7.90\%$.
This indicates that reasoning in culture-associated languages promotes more targeted and relevant knowledge retrieval, rather than broader but noisier recall, providing a concrete mechanistic explanation for the persistent gains observed in cultural tasks.

\section{Language Mixing Behavior}\label{app:langmix}

We present a example of language mixing behavior in Figure~\ref{fig:example_langmix}, where the default reasoning in English is interleaved with reasoning in Japanese.

\section{Further Enhancement with Direct Preference Optimization}\label{app:dpo}
In our main experiments, our trainining strategy only leverages the signals from the advantageous language.
However, the instance-level contrastive signals between the advantageous and less advantageous languages naturally induce implicit preferences between alternative reasoning-language choices.
In this subsection, we further apply Direct Preference Optimization (DPO) as an auxiliary enhancement, treating reasoning trajectories in the advantageous language as preferred samples and those in the less advantageous language as dispreferred ones. The results are shown in Table~\ref{tab:main_dpo}.

\definecolor{up-green}{RGB}{0,120,0}
\definecolor{down-red}{RGB}{255,0,0}
\definecolor{ForestGreen}{RGB}{34,139,34}

\begin{table*}[t]
\small
\renewcommand{\arraystretch}{1.2} 
\setlength{\dashlinedash}{4pt} 
\setlength{\dashlinegap}{2pt}  
  \centering
  \resizebox{1.2\columnwidth}{!}{
    \begin{tabular}{l@{\hskip 4pt}cccc}
    \toprule
    \multirow{3}[4]{*}{\textbf{Models}} & \multicolumn{2}{c}{\textbf{Multilingual Math Reasoning}} & \multicolumn{2}{c}{\textbf{Cultural Reasoning}} \\
    \cmidrule(lr){2-3}
    \cmidrule(lr){4-5}
    & \multicolumn{1}{c}{\textbf{MGSM}} & \multicolumn{1}{c}{\textbf{MT-AIME}} & \multicolumn{1}{c}{\textbf{FORK}} & \multicolumn{1}{c}{\textbf{CulturalBench}} \\
    & \multicolumn{1}{c}{\underline{\textbf{\textit{Think}}}} & \multicolumn{1}{c}{\underline{\textbf{\textit{Think}}}} & \multicolumn{1}{c}{\underline{\textbf{\textit{Think}}}} & \multicolumn{1}{c}{\underline{\textbf{\textit{Think}}}} \\
    \midrule
    \multicolumn{5}{c}{\textbf{\textit{Open-sourced Reasoning Models}}} \\
    \midrule
    \text{Qwen3-4B} & 76.59 & 21.78 & 73.73 & 70.85 \\
    \text{Qwen3-14B} & 82.56  & 29.22 & 73.91 & 78.24 \\
    \text{Qwen3-32B} & 83.98  & 33.89 & 81.88 & 81.26 \\
    \midrule
    \multicolumn{5}{c}{\textbf{\textit{x1 Series Models}}} \\
    \midrule
    \noalign{\vskip -0.8ex}
    & \multicolumn{2}{c}{\textbf{\textit{\cellcolor{cyan!15}+ Math}}} & \multicolumn{2}{c}{\textbf{\textit{\cellcolor{up-green!15}+ Culture}}} \\
    {\textbf{\textit{x1}}-Qwen3-4B}
    & 77.69 & 22.83 & 78.08 & 72.74 \\
    {\textbf{\textit{x1}}-Qwen3-14B}
    & 83.64 & 33.11 & 76.81 & 81.58  \\
    {\textbf{\textit{x1}}-Qwen3-32B}
    & 84.43 & 34.50 & 82.61 &  81.99 \\
    \midrule
    \multicolumn{5}{c}{\textbf{\textit{x1 Series Models + DPO}}} \\
    \midrule
    \noalign{\vskip -0.8ex}
    & \multicolumn{2}{c}{\textbf{\textit{\cellcolor{cyan!15}+ Math}}} & \multicolumn{2}{c}{\textbf{\textit{\cellcolor{up-green!15}+ Culture}}} \\
    {\textbf{\textit{x1}}-Qwen3-4B}
    & 78.23 & 22.00 & 77.72 & 72.51 \\
    {\textbf{\textit{x1}}-Qwen3-14B}
    & 83.56 & 33.33 & 77.90 & 81.34 \\
    {\textbf{\textit{x1}}-Qwen3-32B}
    & 84.56 & 38.67 & 83.70 & 81.05 \\
    \bottomrule
    \end{tabular}
  }
  \caption{Results after applying DPO to \textit{x1} models.}
  \label{tab:main_dpo}
\end{table*}

\begin{table*}[t]
    \small
    \renewcommand{\arraystretch}{1.2} 
    \setlength{\tabcolsep}{4pt}
    \setlength{\dashlinedash}{4pt} 
    \setlength{\dashlinegap}{2pt}  
      \centering
      \resizebox{\textwidth}{!}{
        \begin{tabular}{lccccccccccc}
        \toprule
         & \multicolumn{11}{c}{\textit{MGSM}} \\
        \cmidrule(lr){2-12}
        & Bn & De & En & Es & Fr & Ja & Ru & Sw & Th & Zh & Average \\
        \midrule
        & \multicolumn{11}{c}{\textbf{\textit{Backbone Models}}} \\
        \midrule
        \text{Qwen3-4B (non-think)} & 68.67 & 76.27 & 83.87 & 75.73 & 74.40 & 73.07 & 81.20 & 15.87 & 74.00 & 79.07 & 70.21  \\
        \text{Qwen3-4B (think)} & 75.33 & 82.53 & 87.60 & 84.67 & 81.60 & 78.67 & 84.67 & 26.80 & 79.33 & 84.67 & 76.59 \\
        \cdashline{1-12}\noalign{\vskip 0.4ex}
        \text{Qwen3-14B (non-think)} & 80.20 & 81.20 & 86.00 & 80.60 & 78.40 & 78.60 & 82.80 & 43.80 & 81.60 & 83.20 & 77.64  \\
        \text{Qwen3-14B (think)} & 83.60 & 85.60 & 86.60 & 84.60 & 82.80 & 82.20 & 87.00 & 60.80 & 86.00 & 86.40 & 82.56 \\
        \cdashline{1-12}\noalign{\vskip 0.4ex}
        \text{Qwen3-32B (non-think)} & 80.8 & 80.8 & 85.6 & 82 & 77.6 & 80.8 & 84.4 & 65.2 & 83.6 & 84.4 & 80.52 \\
        \text{Qwen3-32B (think)} & 84.00 & 86.60 & 89.20 & 84.20 & 83.20 & 82.80 & 86.80 & 71.80 & 84.20 & 87.00 & 83.98 \\
        \cdashline{1-12}\noalign{\vskip 0.4ex}
        \text{DeepSeek-R1-Distill-Qwen-7B (non-think)} & 46.00 & 61.00 & 80.40 & 69.00 & 63.40 & 45.60 & 59.20 & 5.00 & 44.20 & 73.80 & 54.76 \\
        \text{DeepSeek-R1-Distill-Qwen-7B (think)} & 50.40 & 67.07 & 80.00 & 71.07 & 68.80 & 51.87 & 71.87 & 7.07 & 52.13 & 80.27 & 60.05 \\
        \cdashline{1-12}\noalign{\vskip 0.4ex}
        \text{DeepSeek-R1-Distill-Llama-8B (non-think)} & 12.13 & 46.00 & 65.73 & 49.60 & 45.60 & 35.33 & 47.33 & 3.73 & 19.87 & 56.40 & 38.17 \\
        \text{DeepSeek-R1-Distill-Llama-8B (think)} & 11.73 & 45.07 & 66.40 & 51.87 & 47.07 & 33.87 & 44.93 & 5.07 & 25.47 & 72.13 & 40.36 \\
        \midrule
        & \multicolumn{11}{c}{\textbf{\textit{x1 Series Models}}} \\
        \midrule
        \text{\textit{x1}-Qwen3-4B (non-think)} & 66.20 & 76.40 & 84.80 & 78.00 & 75.60 & 71.40 & 78.80 & 17.80 & 76.60 & 77.40 & 70.30 \\
        \text{\textit{x1}-Qwen3-4B (think)} & 77.47 & 84.53 & 91.07 & 85.47 & 81.87 & 78.93 & 85.47 & 27.33 & 80.13 & 84.67 & 77.69 \\
        \cdashline{1-12}\noalign{\vskip 0.4ex}
        \text{\textit{x1}-Qwen3-14B (non-think)} & 78.00 & 79.20 & 86.40 & 79.00 & 77.40 & 80.00 & 83.00 & 44.40 & 84.40 & 82.00 & 77.38 \\
        \text{\textit{x1}-Qwen3-14B (think)} & 82.00 & 86.00 & 88.80 & 86.20 & 84.00 & 83.00 & 87.60 & 61.20 & 88.40 & 89.20 & 83.64 \\
        \cdashline{1-12}\noalign{\vskip 0.4ex}
        \text{\textit{x1}-Qwen3-32B (non-think)} & 82.00 & 82.40 & 87.60 & 82.40 & 76.80 & 80.00 & 84.80 & 62.00 & 78.40 & 84.80 & 80.12 \\
        \text{\textit{x1}-Qwen3-32B (think)} & 85.47 & 84.80 & 89.87 & 85.07 & 83.07 & 84.40 & 87.33 & 73.33 & 84.27 & 86.67 & 84.43 \\
        \cdashline{1-12}\noalign{\vskip 0.4ex}
        {\textbf{\textit{x1}}-DeepSeek-R1-Distill-Qwen-7B (non-think)} & 50.00 & 57.20 & 80.40 & 68.00 & 60.80 & 50.80 & 60.40 & 4.80 & 43.60 & 69.20 & 54.52 \\
        {\textbf{\textit{x1}}-DeepSeek-R1-Distill-Qwen-7B (think)} & 57.60 & 71.40 & 80.20 & 72.40 & 67.80 & 60.40 & 74.00 & 9.80 & 59.00 & 79.80 & 63.24 \\
        \cdashline{1-12}\noalign{\vskip 0.4ex}
        {\textbf{\textit{x1}}-DeepSeek-R1-Distill-Llama-8B (non-think)} & 10.67 & 46.00 & 65.60 & 49.60 & 45.60 & 35.33 & 47.33 & 3.73 & 19.87 & 56.40 & 38.01 \\
        {\textbf{\textit{x1}}-DeepSeek-R1-Distill-Llama-8B (think)} & 26.40 & 60.53 & 76.53 & 62.80 & 62.67 & 49.73 & 64.93 & 12.93 & 37.20 & 68.00 & 52.17 \\
        \bottomrule
        & \multicolumn{11}{c}{\textit{MT-AIME}} \\
        \cmidrule(lr){2-12}
        & Bn & De & En & Es & Fr & Ja & Ru & Sw & Th & Zh & Average \\
        \midrule
        & \multicolumn{11}{c}{\textbf{\textit{Backbone Models}}} \\
        \midrule
        \text{Qwen3-4B (non-think)} & 6.67 & 16.67 & 17.78 & 12.22 & 23.33 & 7.78 & 15.56 & 4.44 & 7.78 & 16.67 & 12.89  \\
        \text{Qwen3-4B (think)} & 22.22 & 25.56 & 25.56 & 24.44 & 23.33 & 22.22 & 17.78 & 6.67 & 25.56 & 24.44 & 21.78 \\
        \cdashline{1-12}\noalign{\vskip 0.4ex}
        \text{Qwen3-14B (non-think)} & 16.67 & 26.67 & 25.00 & 25.00 & 21.67 & 18.33 & 21.67 & 5.00 & 13.33 & 20.00 & 19.33  \\
        \text{Qwen3-14B (think)} & 26.67 & 26.67 & 32.22 & 25.56 & 34.44 & 24.44 & 34.44 & 18.89 & 23.33 & 45.56 & 29.22 \\
        \cdashline{1-12}\noalign{\vskip 0.4ex}
        \text{Qwen3-32B (non-think)} & 23.33 & 21.67 & 26.67 & 21.67 & 21.67 & 25.00 & 26.67 & 10.00 & 21.67 & 20.00 & 21.83 \\
        \text{Qwen3-32B (think)} & 28.89 & 28.89 & 26.67 & 43.33 & 31.11 & 31.11 & 38.89 & 27.78 & 35.56 & 46.67 & 33.89 \\
        \cdashline{1-12}\noalign{\vskip 0.4ex}
        \text{DeepSeek-R1-Distill-Qwen-7B (non-think)} & 2.22 & 6.67 & 16.67 & 11.11 & 7.78 & 3.33 & 10.00 & 3.33 & 6.67 & 15.56 & 8.33 \\
        \text{DeepSeek-R1-Distill-Qwen-7B (think)} & 23.33 & 31.67 & 33.33 & 25.00 & 30.00 & 28.33 & 26.67 & 8.33 & 20.00 & 31.67 & 25.83 \\
        \cdashline{1-12}\noalign{\vskip 0.4ex}
        \text{DeepSeek-R1-Distill-Llama-8B (non-think)} & 1.11 & 2.22 & 1.11 & 4.44 & 3.33 & 1.11 & 3.33 & 0.00 & 4.44 & 5.56 & 2.67 \\
        \text{DeepSeek-R1-Distill-Llama-8B (think)} & 11.11 & 16.67 & 15.56 & 18.89 & 17.78 & 12.22 & 15.56 & 5.56 & 11.11 & 20.00 & 14.44 \\
        \midrule
        & \multicolumn{11}{c}{\textbf{\textit{x1 Series Models}}} \\
        \midrule
        \text{\textit{x1}-Qwen3-4B (non-think)} &  6.67 & 18.89 & 23.33 & 15.56 & 16.67 & 8.89 & 15.56 & 5.56 & 7.78 & 16.67 & 13.56 \\
        \text{\textit{x1}-Qwen3-4B (think)} & 16.67 & 25.00 & 25.00 & 25.00 & 23.33 & 21.67 & 23.33 & 10.00 & 25.00 & 33.33 & 22.83  \\
        \cdashline{1-12}\noalign{\vskip 0.4ex}
        \text{\textit{x1}-Qwen3-14B (non-think)} & 14.44 & 21.11 & 23.33 & 23.33 & 24.44 & 17.78 & 27.78 & 7.78 & 18.89 & 15.56 & 19.44 \\
        \text{\textit{x1}-Qwen3-14B (think)} & 20.00 & 34.44 & 38.89 & 35.56 & 27.78 & 36.67 & 32.22 & 20.00 & 33.33 & 52.22 & 33.11 \\
        \cdashline{1-12}\noalign{\vskip 0.4ex}
        \text{\textit{x1}-Qwen3-32B (non-think)} & 18.89 & 21.11 & 24.44 & 25.56 & 16.67 & 23.33 & 22.22 & 18.89 & 25.56 & 24.44 & 22.11 \\
        \text{\textit{x1}-Qwen3-32B (think)} &  31.67 & 35.00 & 35.00 & 33.33 & 36.67 & 40.00 & 35.00 & 23.33 & 30.00 & 45.00 & 34.50 \\
        \cdashline{1-12}\noalign{\vskip 0.4ex}
        {\textbf{\textit{x1}}-DeepSeek-R1-Distill-Qwen-7B (non-think)} & 7.78 & 8.89 & 12.22 & 11.11 & 7.78 & 4.44 & 10.00 & 6.67 & 10.00 & 11.11 & 9.00 \\
        {\textbf{\textit{x1}}-DeepSeek-R1-Distill-Qwen-7B (think)} & 23.33 & 26.67 & 23.33 & 26.67 & 30.00 & 23.33 & 26.67 & 23.33 & 33.33 & 33.33 & 27.00  \\
        \cdashline{1-12}\noalign{\vskip 0.4ex}
        {\textbf{\textit{x1}}-DeepSeek-R1-Distill-Llama-8B (non-think)} & 2.22 & 3.33 & 3.33 & 2.22 & 2.22 & 3.33 & 2.22 & 0.00 & 2.22 & 7.78 & 2.89 \\
        {\textbf{\textit{x1}}-DeepSeek-R1-Distill-Llama-8B (think)} & 3.33 & 23.33 & 23.33 & 20.00 & 23.33 & 20.00 & 16.67 & 3.33 & 16.67 & 20.00 & 17.00  \\
        \bottomrule
        \end{tabular}
      }
      \caption{Results for each subsets in \textit{MGSM} and \textit{MT-AIME} across different languages.}
      \label{tab:main_math}
    \end{table*}

\end{document}